\documentclass{article}
\pdfoutput=1
\usepackage[final,nonatbib]{neurips_2020}

\usepackage{microtype}
\usepackage{graphicx}
\usepackage{booktabs}
\usepackage{amsmath}
\usepackage{amsthm}
\usepackage{amssymb}
\usepackage{mathtools}
\usepackage{subcaption}
\usepackage{algorithm}
\usepackage{algorithmic}

\usepackage[square,sort,comma,numbers]{natbib}
\usepackage{hyperref}

\DeclarePairedDelimiter\round{\lfloor}{\rceil}

\title{Bayesian Bits: Unifying Quantization and Pruning}

\author{%
  Mart van Baalen\thanks{Equal contribution}, Christos Louizos\footnotemark[1], Markus Nagel, Rana Ali Amjad, \\ 
  {\bf Ying Wang, Tijmen Blankevoort, Max Welling}\\
  Qualcomm AI Research\thanks{Qualcomm AI Research is an initiative of Qualcomm Technologies, Inc.}\\
  \texttt{\{mart,clouizos,markusn,ramjad,yinwan,tijmen,mwelling\}@qti.qualcomm.com} \\
}

\begin{document}

\maketitle

\begin{abstract}
  We introduce Bayesian Bits, a practical method for joint mixed precision quantization and pruning through gradient based optimization.
Bayesian Bits employs a novel decomposition of the quantization operation, which sequentially considers doubling the bit width.
At each new bit width, the residual error between the full precision value and the previously rounded value is quantized.
We then decide whether or not to add this quantized residual error for a higher effective bit width and lower quantization noise.
By starting with a power-of-two bit width, this decomposition will always produce hardware-friendly configurations, and through an additional 0-bit option, serves as a unified view of pruning and quantization.
Bayesian Bits then introduces learnable stochastic gates, which collectively control the bit width of the given tensor. 
As a result, we can obtain low bit solutions by performing approximate inference over the gates, with prior distributions that encourage most of them to be switched off. 
We experimentally validate our proposed method on several benchmark datasets and show that we can learn pruned, mixed precision networks that provide a better trade-off between accuracy and efficiency than their static bit width equivalents.
\end{abstract}

\section{Introduction}\label{sec:introduction}
To reduce the computational cost of neural network inference, quantization and compression techniques are often applied before deploying a model in real life.
The former reduces the bit width of weight and activation tensors by quantizing floating-point values onto a regular grid, allowing the use of cheap integer arithmetic, while the latter aims to reduce the total number of multiply-accumulate (MAC) operations required.
We refer the reader to \cite{krishnamoorthi2018quantizing} and \cite{kuzmin2019taxonomy} for overviews of hardware-friendly quantization and compression techniques, respectively.

In quantization, the default assumption is that all layers should be quantized to the same bit width.
While it has long been understood that low bit width quantization can be achieved by keeping the first and last layers of a network in higher precision \citep{shayer2017learning,choi2018pact}, recent work \citep{dong2019hawq,uhlich2019dq,wang2019haq} has shown that carefully selecting the bit width of each tensor can yield a better trade-off between accuracy and complexity.
Since the choice of quantization bit width for one tensor may affect the quantization sensitivity of all other tensors, the choice of bit width cannot be made without regarding the rest of the network.

The number of possible bit width configurations for a neural network is exponential in the number of layers in the network.
Therefore, we cannot exhaustively search all possible configurations and pick the best one.
Several approaches to learning the quantization bit widths from data have been proposed, either during training \citep{uhlich2019dq,louizos2017bayesian}, or on pre-trained networks \citep{wang2019haq,dong2019hawq,dong2019hawqv2}.
However, these works do not consider the fact that commercially available hardware typically only supports efficient computation in power-of-two bit widths (see, e.g., \cite{ignatov2019aibenchmark} for a mobile hardware overview and \cite{moons2017isscc} for a method to perform four 4-bit multiplications in a 16-bit hardware multiplication unit.)

In this paper, we introduce a novel decomposition of the quantization operation.
This decomposition exposes all hardware-friendly (i.e., power-of-two) bit widths individually by recursively quantizing the residual error of lower bit width quantization.
The quantized residual error tensors are then added together into a quantized approximation of the original tensor. This allows for the introduction of learnable gates: by placing a gate on each of the quantized residual error tensors, the effective bit width can be controlled, thus allowing for data-dependent optimization of the bit width of each tensor, which we learn jointly with the (quantization) scales and network parameters.
We then extend the gating formulation such that not only the residuals, but the overall result of the quantization is gated as well. This facilitates for ``zero bit'' quantization and serves as a unified view of pruning and quantization.
We cast the optimization of said gates as a variational inference problem with prior distributions that favor quantizers with low bit widths. 
Lastly, we provide an intuitive and practical approximation to this objective, that is amenable to efficient gradient-based optimization.
We experimentally validate our method on several models and datasets and show encouraging results, both for end-to-end fine-tuning tasks as well as post-training quantization.

\vspace{-0.2cm}
\section{Unifying quantization and pruning with Bayesian Bits}\label{sec:method}
\vspace{-0.2cm}
Consider having an input $x$ in the range of $[\alpha, \beta]$ that is quantized with a uniform quantizer with an associated bit width $b$. Such a quantizer can be expressed as
\begin{align}
    x_q = s \round{x / s}, \qquad s = \frac{\beta - \alpha}{2^b - 1}
\end{align}
where $x_q$ is a quantized approximation of $x$, $\round{\cdot}$ indicates the round-to-nearest-integer function, and $s$ is the step-size of the quantizer that depends on the given bit width $b$. 
How can we learn the number of bits $b$, while respecting the hardware constraint that $b$ should be a power of two? One possible way would be via ``decomposing'' the quantization operation in a way that exposes all of the appropriate bit widths. In the following section, we will devise a simple and practical method that realizes such a procedure.

\subsection{Mixed precision gating for quantization and pruning}
\begin{figure}[t]
\begin{centering}
  \includegraphics[width=\linewidth]{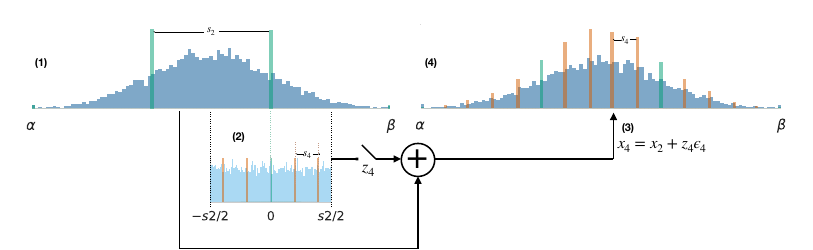}
  \caption{\textbf{Illustration of our decomposition.}  The input floating point values $x$ are clipped to the learned range $[\alpha, \beta]$ (dark blue histograms), and are quantized to 2 bits into $x_2$ (green histograms) (1). 
  To accommodate the $2^2$ grid points of the 2 bit quantization grid, the range is divided into $2^2-1$ equal parts, hence $s_2=\frac{\beta-\alpha}{2^2-1}$. 
  Next, the residual error $x-x_2$ is computed (light blue histogram), and quantized onto the 4 bit grid (2), resulting in the quantized residual error tensor $\epsilon_4$.
  To accommodate the points of the 4 bit quantization grid, the range is divided into $2^4-1$ equal parts. 
  Note that $(2^4-1)=(2^2-1)(2^2+1)$, thus we can compute $s_4$ as $s_2 / (2^2+1)$.
  This can alternatively be seen as dividing the residual error, with range bounded by $[-s_2/2, s_2/2]$, into $2^2+1$ equal parts.
  Values in the quantized residual error equal to 0 correspond to points on the 2 bit grid, other values correspond to points on the 4 bit grid (orange histogram).
  Next, the quantized residual error is added to $x_2$ if the 4-bit gate $z_4$ is equal to 1 (3), resulting in the 4-bit quantized tensor $x_4$ (4).
  NB: quantization histograms and floating point histograms are not on the same scale.
  }
  \label{fig:decomposition}\vspace{-0.5cm}
\end{centering}
\end{figure}

Consider initially quantizing $x$ with $b=2$:
\begin{align}
    x_2 = s_2 \round{x / s_2}, \qquad s_2 = \frac{\beta - \alpha}{2^2 - 1}.
\end{align}
How can we then ``move'' to the next hardware friendly bit width, i.e., $b=4$? We know that the quantization error of this operation will be $x - x_2$, and it will be in $[-s_2 / 2, s_2 / 2]$. 
We can then consider encoding this residual error according to a fixed point grid that has a length of $s_2$ and bins of size $s_2 / (2^2 + 1)$
\begin{align}
    \epsilon_4 = s_4 \round{(x - x_2) / s_4}, \qquad s_4 = \frac{s_2}{2^2 + 1}.
\end{align}
By then adding this quantized residual to $x_2$, i.e. $x_4 = x_2 + \epsilon_4$ we obtain a quantized tensor $x_4$ that has double the precision of the previous tensor, i.e. an effective bit width of $b=4$ with a step-size of $s_4 = \frac{\beta - \alpha}{(2^2 - 1) (2^2 + 1)} = \frac{\beta - \alpha}{2^4 - 1}$. To understand why this is the case, we can proceed as follows: the output of $s_2 \round{x / s_2}$ will be an integer multiple of $s_4$, as $s_2 = s_4 (2^2 + 1)$, thus it will be a part of the four bit quantization grid as well. Furthermore, the quantized residual is also an integer multiple of $s_4$, as $\round{(x - x_2) / s_4}$ produces elements in $\{-2, -1, 0, 1, 2\},$ thus it corresponds to a simple re-assignment of $x$ to a different point on the four bit grid.
See Figure~\ref{fig:decomposition} for an illustration of this decomposition.

This idea can be generalized to arbitrary power of two bit widths by sequentially doubling the precision of the quantized tensor through the addition of the, quantized, remaining residual error
\begin{align}
    x_q & = x_2 + \epsilon_4 + \epsilon_8 + \epsilon_{16} + \epsilon_{32}
\end{align}
where each quantized residual is $\epsilon_b = s_b \round{(x - x_{b/2}) / s_b}$, with a step size $s_b = s_{b/2} / (2^{b/2} + 1)$, and previously quantized value $x_{b/2} = x_2 + \sum_{2< j \leq b/2}\epsilon_j$ for $b \in \{4, 8, 16, 32\}$. In this specific example, $x_q$ will be quantized according to a 32-bit fixed point grid. Our lowest bit width is 2-bit to allow for the representation of 0, e.g. in the case of padding in convolutional layers.

Having obtained this decomposition, we then seek to learn the appropriate bit width. 
We introduce gating variables $z_i$, i.e. variables that take values in $\{0, 1\}$, for each residual error  $\epsilon_i$. More specifically, we can express the quantized value as
\begin{align}
    x_q = x_2  + z_4 (\epsilon_4 + z_8 (\epsilon_8 + z_{16} (\epsilon_{16} + z_{32} \epsilon_{32}))). \label{eq:decomp_quant}
\end{align}
If one of the gates $z_i$ takes the value of zero, it completely de-activates the addition of all of the higher bit width residuals, thus controlling the effective bit width of the quantized value $x_q$. Actually, 
we can take this a step further and consider pruning as quantization with a zero bit width. We can thus extend Eq.~\ref{eq:decomp_quant} as follows:
\begin{align}
    x_q = z_2 (x_2  + z_4 (\epsilon_4 + z_8 (\epsilon_8 + z_{16} (\epsilon_{16} + z_{32} \epsilon_{32})))), \label{eq:decomp_prun_quant}
\end{align}
where now we also introduce a gate for the lowest bit width possible, $z_2$. 
If that particular gate is switched off, then the input $x$ is  assigned the value of $0$, thus quantized to 0-bit and pruned away. 
Armed with this modification, we can then perform, e.g., structured pruning by employing a separate quantizer of this form for each filter in a convolutional layer. To ensure that the elements of the tensor that survive the pruning will be quantized according to the same grid, we can share the gating variables for $b > 2$, along with the quantization grid step sizes. 

\subsection{Bayesian Bits}
We showed in Eq.~\ref{eq:decomp_prun_quant} that quantizing to a specific bit width can be seen as a gated addition of quantized residuals. We want to incorporate a principled regularizer for the gates, such that it encourages gate configurations that prefer efficient neural networks. We also want a learning algorithm that allows us to apply efficient gradient based optimization for the binary gates $z$, which is not possible by directly considering Eq.~\ref{eq:decomp_prun_quant}. We show how to tackle both issues through the lens of Bayesian, and more specifically, variational inference; we derive a gate regularizer through a prior that favors low bit width configurations and a learning mechanism that allows for gradient based optimization. 

For simplicity, let us assume that we are working on a supervised learning problem, where we are provided with a dataset of $N$ i.i.d. input-output pairs $\mathcal{D} = \{(\mathbf{x}_i, y_i)\}_{i=1}^{N}$. Furthermore, let us assume that we have a neural network with parameters $\theta$ and a total of $K$ quantizers that quantize up to 8-bit\footnote{This is just for simplifying the exposition and not a limitation of our method.} with associated gates $\mathbf{z}_{1:K}$, where $\mathbf{z}_i = [z_{2i}, z_{4i}, z_{8i}]$. We can then use the neural network for the conditional distribution of the targets given the inputs, i.e. $p_\theta(\mathcal{D}|\mathbf{z}_{1:K}) = \prod_{i=1}^{N} p_\theta(y_i|x_i, \mathbf{z}_{1:K})$. Consider also positing a prior distribution (which we will discuss later) over the gates $p(\mathbf{z}_{1:K}) = \prod_{k}p(\mathbf{z}_k)$. We can then perform variational inference with an approximate posterior that has parameters $\phi$, $q_\phi(\mathbf{z}_{1:K}) = \prod_{k}q_\phi(\mathbf{z}_k)$ by maximizing the following lower bound to the marginal likelihood $p_\theta(\mathcal{D})$~\citep{peterson1987mean,hinton1993keeping}
\begin{align}
     \mathcal{L}(\theta, \phi) =  \mathbb{E}_{q_\phi(\mathbf{z}_{1:K})}[\log p_\theta(\mathcal{D}|\mathbf{z}_{1:K})] - \sum_k KL(q_\phi(\mathbf{z}_k) || p(\mathbf{z}_k)). \label{eq:elbo_bb}
\end{align}
The first term can be understood as the ``reconstruction'' term, which aims to obtain good predictive performance for the targets given the inputs. The second term is the ``complexity'' term that, through the KL divergence, aims to regularize the variational posterior distribution to be as close as possible to the prior $p(\mathbf{z}_{1:K})$. 
Since each addition of the quantized residual doubles the bit width, let us assume that the gates $\mathbf{z}_{1:K}$ are binary; we either double the precision of each quantizer or we keep it the same. 
We can then set up an autoregressive prior and variational posterior distribution for the next bit configuration of each quantizer $k$, conditioned on the previous, as follows:
\begin{align}
    &p(z_{2k}) = \text{Bern}(e^{- \lambda}), \enskip q_\phi(z_{2k}) = \text{Bern}(\sigma(\phi_{2k})),\\
    &p(z_{4k} | z_{2k} = 1) = p(z_{8k} | z_{4k} = 1) = \text{Bern}(e^{- \lambda}), \\
    &q_\phi(z_{4k}| z_{2k} = 1) = \text{Bern}(\sigma(\phi_{4k})), \enskip q_\phi(z_{8k}| z_{4k} = 1) = \text{Bern}(\sigma(\phi_{8k}))\\
    & p(z_{4k} | z_{2k} = 0) = p(z_{8k}| z_{4k} = 0) = \text{Bern}(0),\\
    & q(z_{4k} | z_{2k} = 0) = q(z_{8k} | z_{4k} = 0) = \text{Bern}(0),
\end{align}
where $e^{-\lambda}$ with $\lambda \geq 0$ is the prior probability of success and $\sigma(\phi_{ik})$ is the posterior probability of success with sigmoid function $\sigma(\cdot)$ and $\phi_{ik}$ the learnable parameters. This structure encodes the fact that when the gate for e.g. 4-bit is ``switched off'', the gate for 8-bit will also be off. 
For brevity, we will refer to the variational distribution that conditions on an active previous bit as $q_\phi(z_{ik})$ instead of $q_\phi(z_{ik} | z_{i/2,k} = 1)$, since the ones conditioned on a previously inactive bit, $q_\phi(z_{ik} | z_{i/2,k} = 0)$, are fixed. 
The KL divergence for each quantizer in the variational objective then decomposes to:
\begin{align}
    &KL(q_\phi(\mathbf{z}_k) || p(\mathbf{z}_k)) = KL(q_\phi(z_{2k}) || p(z_{2k})) + q_\phi(z_{2k} = 1) KL(q_\phi(z_{4k}) || p(z_{4k}| z_{2k} = 1)) + \nonumber \\
    & \enskip q_\phi(z_{2k} = 1)q_\phi(z_{4k} = 1)KL(q_\phi(z_{8k}) || p(z_{8k}| z_{4k} = 1))\label{eq:kl_autoregressive}
\end{align}
We can see that the posterior inclusion probabilities of the lower bit widths downscale the KL divergence of the higher bit widths. This is important, as the gates for the higher order bit widths can only contribute to the log-likelihood of the data when the lower ones are active due to their multiplicative interaction. Therefore, the KL divergence at Eq.~\ref{eq:kl_autoregressive} prevents the over-regularization that would have happened if we had assumed fully factorized distributions.

\subsection{A simple approximation for learning the bit width}
So far we have kept the prior as an arbitrary Bernoulli with a specific form for the probability of inclusion, $e^{- \lambda}$. How can we then enforce that the variational posterior will ``prune away'' as many gates as possible? The straightforward answer would be by choosing large values for $\lambda$; for example, if we are interested in networks that have low computational complexity, we can set $\lambda$ proportional to the Bit Operation (BOP) count contribution of the particular object that is to be quantized. By writing out the KL divergence with this specific prior for a given KL term, we will have that
\begin{align}
    KL(q_\phi(z_{ik})) || & p(z_{ik})) = - H[q_\phi] + \lambda q(z_{ik} = 1) - \log (1 - e^{- \lambda})(1 - q(z_{ik} = 1)),
\end{align}
where $H[q_\phi]$ corresponds to the entropy of the variational posterior $q_\phi(z_{ik})$. Now, under the assumption that $\lambda$ is sufficiently large, we have that $(1 - e^{- \lambda}) \approx 1$, thus the third term of the r.h.s. vanishes. 
Furthermore, let us assume that we want to optimize a rescaled version of the objective at Eq.~\ref{eq:elbo_bb} where, without changing the optima, we divide both the log-likelihood and the KL-divergence by the size of the dataset $N$.
In this case the individual KL divergences will be
\begin{align}
    \frac{1}{N}KL(q_\phi(z_{ik}) || p(z_{ik})) \approx  - \frac{1}{N}H[q_\phi] + \frac{\lambda}{N} q_\phi(z_{ik} = 1).
\end{align}
For large $N$ the contribution of the entropy term will then be negligible. Equivalently, we can consider doing MAP estimation on the objective of Eq.~\ref{eq:elbo_bb}, which corresponds to simply ignoring the entropy terms of the variational bound. 
Now consider scaling the prior with $N$, i.e. $\lambda = N \lambda^\prime$. This denotes that the number of gates that stay active is constant irrespective of the size of the dataset. As a result, whereas for large $N$ the entropy term is negligible the contribution from the prior is still significant. 
Thus, putting everything together, we arrive at a simple and intuitive objective function 
\begin{equation}
    \mathcal{F}(\theta, \phi) := \mathbb{E}_{q_\phi(\mathbf{z}_{1:K})}\left[\frac{1}{N}\log p_\theta(\mathcal{D}|\mathbf{z}_{1:K})\right] - \lambda^\prime \sum_{k}\sum_{i \in B}\prod_{j \in B}^{j \leq i} q_\phi(z_{jk} = 1), \label{eq:bb_map}
\end{equation}
where $B$ corresponds to the available bit widths of the quantizers.
 This objective can be understood as penalizing the probability of including the set of parameters associated with each quantizer and additional bits of precision assigned to them. The final objective reminisces the $L_0$ norm regularization from~\cite{louizos2018learning}; indeed, under some assumptions in Bayesian Bits we recover the same objective. We discuss the relations between those two algorithms further in the Appendix.
 
\subsection{Practical considerations}
The final objective we arrived at in Eq.~\ref{eq:bb_map} requires us to compute an expectation of the log-likelihood with respect to the stochastic gates. For a moderate amount of gates, this can be expensive to compute. One straightforward way to avoid it is to approximate the expectation with a Monte Carlo average by sampling from $q_\phi(\mathbf{z}_{1:K})$ and using the REINFORCE estimator~\cite{williams1992simple}. While this is straightforward to do, the gradients have high variance which, empirically, hampers the performance. 
To obtain a better gradient estimator with lower variance we exploit the connection of Bayesian Bits to $L_0$ regularization and employ the hard-concrete relaxations of~\cite{louizos2018learning} as $q_\phi(\mathbf{z}_{1:K})$, 
thus allowing for gradient-based optimization through the reparametrization trick~\citep{kingma2013auto, rezende2014stochastic}. 
At test time, the authors of~\cite{louizos2018learning} propose a deterministic variant of the gates where the noise is switched off. As that can result into gates that are not in $\{0,1\}$, thus not exactly corresponding to doubling the bits of precision, we take an alternative approach. We prune a gate whenever the probability of exact $0$ under the relaxation  exceeds a threshold $t$, otherwise we set it to $1$.  One could also hypothesize alternative ways to learn the gates, but we found that other approaches yielded inferior results.
We provide all of the details about the Bayesian Bits optimization, test-time thresholding and alternative gating approaches in the Appendix.

For the decomposition of the quantization operation that we previously described, we also need the inputs to be constrained within the quantization grid $[\alpha, \beta]$. A simple way to do this would be to clip the inputs before pushing them through the quantizer. For this clipping we will use PACT~\citep{choi2018pact}, which in our case clips the inputs according to
\begin{align}
    &\text{clip}(x; \alpha, \beta) = \beta - \text{ReLU}(\beta - \alpha - \text{ReLU}(x - \alpha))\label{eq:pact_clip}
\end{align}
where $\beta, \alpha$ can be trainable parameters. In practice we only learn $\beta$ as we set $\alpha$ to zero for unsigned quantization (e.g. for ReLU activations), and for signed quantization we set $\alpha = -\beta$. We  subtract a small epsilon from $\beta$ via $(1 - 10^{-7})\beta$ before we use it at Eq.~\ref{eq:pact_clip}, to ensure that we avoid the corner case in which a value of exactly $\beta$ is rounded up to an invalid grid point. The step size of the initial grid is then parametrized as $s_2 = \frac{\beta - \alpha}{2^2 - 1}$.

Finally, for the gradients of the network parameters $\theta$, we follow the standard practice and employ the straight-through estimator (STE)~\citep{bengio2013estimating} for the rounding operation, i.e., we perform the rounding in the forward pass but ignore it in the backward pass by assuming that the operation is the identity.

\vspace{-0.2cm}
\section{Related work}\label{sec:backgroundrelated}
\vspace{-0.2cm}
The method most closely related to our work is Differentiable Quantization~(DQ)~\citep{uhlich2019dq}. 
In this method, the quantization range and scale are learned from data jointly with the model weights, from which the bit width can be inferred.
However, for a hardware-friendly application of this method, the learned bit widths must be rounded up to the nearest power-of-two.
As a result, hypothetical efficiency gains will likely not be met in reality.
Several other methods for finding mixed precision configurations have been introduced in the literature. 
\cite{dong2019hawq} and follow-up work~\cite{dong2019hawqv2} use respectively the largest eigenvalue and the trace of the Hessian to determine a layer's sensitivity to perturbations. 
The intuition is that strong curvature at the loss minimum implies that small changes to the weights will have a big impact on the loss. 
Similarly to this work, \cite{louizos2017bayesian} takes a Bayesian approach and determines the bit width for each weight tensor through a heuristic based on the weight uncertainty in the variational posterior. 
The drawback, similarly to~\cite{uhlich2019dq}, of such an approach is that there is no inherent control over the resulting bit widths.

\cite{wu2018mixed} frames the mixed precision search problem as an architecture search.
For each layer in their network, the authors maintain a separate weight tensor for each bit width under consideration.
A stochastic version of DARTS \cite{liu2018darts} is then used to learn the optimal bit width setting jointly with the network's weights.
\cite{wang2019haq} model the assignment of bit widths as a reinforcement learning problem.
Their agent's observation consists of properties of the current layer, and its action space is the possible bit widths for a layer.
The agent receives the validation set accuracy after a short period of fine-tuning as a reward.
Besides the reward, the agent receives direct hardware feedback from a target device. This feedback allows the agent to adapt to specific hardware directly, instead of relying on proxy measures.

Learning the scale along with the model parameters for a fixed bit width network was independently introduced by \cite{esser2019lsq} and \cite{jain2019tuq}. 
Both papers redefine the quantization operation to expose the scale parameter to the learning process, which is then optimized jointly with the network's parameters.
Similarly, \cite{choi2018pact} reformulate the clipping operation such that the range of activations in a network can be learned from data, leading to activation ranges that are more amenable to quantization.

The recursive decomposition introduced in this paper shares similarities with previous work on residual vector quantization \citep{chen2010approximate}, in which the residual error of vectors quantized using K-means is itself (recursively) quantized.
\cite{gong2014compressing} apply this method to neural network weight compression: the size of a network can be significantly reduced by only storing the centroids of K-means quantized vectors. 
Our decomposition also shares similarites with \cite{residualquantization}. 
A crucial difference is that Bayesian bits produces valid fixed-point tensors by construction whereas for \cite{residualquantization} this is not the case.
Concurrent work \cite{zhang2020precision} takes a similar approach to ours. 
The authors restrict themselves to what is essentially one step of our decomposition (without handling the scales), and to conditional gating during inference on activation tensors. The decomposition is not extended to multiple bit widths.

\vspace{-0.2cm}
\section{Experiments}\label{sec:experiments}
\vspace{-0.2cm}
To evaluate our proposed method we conduct experiments on image classification tasks. In every model, we quantized all of the weights and activations (besides the output logits) using per-tensor quantization, and handled the batch norm layers as discussed in~\cite{krishnamoorthi2018quantizing}. We initialized the parameters of the gates to a large value so that the model initially uses its full 32-bit capacity without pruning. 

We evaluate our method on two axes: test set classification accuracy, and bit operations (BOPs), as a hardware-agnostic proxy to model complexity. 
Intuitively the BOP count measures the number of multiplication operations multiplied by the bit widths of the operands. 
To compute the BOP count we use the formula introduced by \cite{baskin2018uniq}, but ignore the terms corresponding to addition in the accumulator since its bit width is commonly fixed regardless of operand bit width. 
We refer the reader to the Appendix for details.
We include pruning by performing group sparsity on the output channels of the weight tensors only, as pruning an output channel of the weight tensor corresponds to pruning that specific activation. 
Output channel group sparsity can often be exploited by hardware \cite{he2017channel}.

Finally, we set the prior probability $p(z_{jk}=1\mid z_{(j/2)k}=1)=e^{-\mu\lambda_{jk}}$, where $\lambda_{jk}$ is proportional to the contribution of gate $z_{jk}$ to the total model BOPs, which is a function of both the tensor $k$ and the bit width $j$, and $\mu$ is a (positive) global regularization parameter. See the Appendix for details.
It is worth noting that improvements in BOP count may not directly correspond to reduced latency on specific hardware. 
Instead, these results should be interpreted as an indication that our method can optimize towards a hardware-like target. 
One could alternatively encourage low memory networks by e.g. using the regularizer from~\citep{uhlich2019dq} or even allow for hardware aware pruning and quantization by using e.g. latency timings from a hardware simulator, similar to \cite{wang2019haq}.

We compare the results of our proposed approach to literature that considers both static as well as mixed precision architectures. 
If BOP counts for a specific model are not provided by the original papers, we perform our own BOP computations, and in some cases we run our own baselines to allow for apples-to-apples comparison to alternative methods (details in Appendix).
All tensors (weight and activation) in our networks are quantized, and the bit widths of all quantizers in our network are learned, contrary to common practice in  literature to keep the first and last layers of the networks in a higher bit width (e.g. \cite{choi2018pact,wu2018mixed}).

Finally, while our proposed method facilitates an end-to-end gradient based optimization for pruning and quantization, in practical applications one might not have access to large datasets and the appropriate compute. 
For this reason, we perform a series of experiments on a consumer-grade GPU using a small dataset, in which only the quantization parameters are updated on a pre-trained model, while the pre-trained weights are kept fixed. 

\subsection{Toy experiments on MNIST \& CIFAR 10}
For the first experiment, we considered the toy tasks of MNIST and CIFAR 10 classification using a LeNet-5 and a VGG-7 model, respectively, commonly employed in the quantization literature, e.g., ~\cite{li2016ternary}. We provide the experimental details in the Appendix. For the CIFAR 10 experiment, we also implemented the DQ method from~\cite{uhlich2019dq} with a BOP regularizer instead of a weight size regularizer so that results can directly be compared to Bayesian Bits. We considered two cases for DQ: one where the bit widths are unconstrained and one where we round up to the nearest bit width that is a power of two (DQ-restricted).

\begin{table*}[t]
    \centering
    \resizebox{\linewidth}{!}{%
    \begin{tabular}{lccccc}
        \toprule %
         & &  \multicolumn{2}{c}{MNIST} & \multicolumn{2}{c}{CIFAR10} \\
         Method & \# bits W/A & Acc. (\%) & Rel. GBOPs (\%)  &  Acc. (\%) & Rel. GBOPs (\%) \\
         \midrule %
         FP32 & 32/32 & 99.36 & 100 & 93.05 & 100 \\
         TWN & 2/32 & 99.35 & 5.74 & 92.56 & 6.22 \\
         LR-Net & 1/32 & 99.47 & 2.99 & 93.18 & 3.11 \\
         RQ & 8/8 & - & - & 93.80 & 6.25 \\
         RQ & 4/4 & - & - & 92.04 & 1.56 \\
         RQ & 2/8 & 99.37 & 0.52 & - & - \\
         WAGE & 2/8 & 99.60 & 1.56 & 93.22 & 1.56 \\
         DQ* & Mixed & - & - & 91.59 & 0.48 \\
         DQ - restricted* & Mixed & - & - & 91.59 & 0.54 \\
         \midrule %
         Bayesian Bits $\mu=0.01$ & Mixed & 99.30$\pm$0.03 & 0.36$\pm$0.01 & 93.23$\pm$0.10 & 0.51$\pm$0.03 \\
         Bayesian Bits $\mu=0.1$ & Mixed & - & - & 91.96$\pm$0.04 & 0.29$\pm$0.00 \\
         \bottomrule
    \end{tabular}
    }
    \caption{Results on the MNIST and CIFAR 10 tasks, mean and stderr over 3 runs. We compare against TWN~\citep{li2016ternary}, LR-Net~\citep{shayer2017learning}, RQ~\citep{louizos2018relaxed}, WAGE~\citep{wu2018training}, and DQ~\citep{uhlich2019dq}. * results run by the authors.
    }\vspace{-0.3cm}
    \label{tab:res_mnist_cifar}
\end{table*}

As we can see from the results in Table~\ref{tab:res_mnist_cifar}, our proposed method provides better trade-offs between the computational complexity of the resulting architecture and the final accuracy on the test set than the baselines which we compare against, both for the MNIST and the CIFAR 10 experiments.
In results for the CIFAR 10 experiments we see that varying the regularization strength can be used to control the trade-off between accuracy and complexity: stronger regularization yields lower accuracy, but also a less complex model.

In the Appendix we plot the learned sparsity and bit widths for our models. 
There we observe that in the aggressive regularization regimes, Bayesian Bits quantizes almost all of the tensors to 2-bit, but usually keeps the first and last layers to higher bit-precision, which is in line with common practice in literature. In the case of moderate regularization at VGG, we observe that Bayesian Bits hardly prunes, it removed 2 channels in the last 256 output convolutional layer and 8 channels at the penultimate weight tensor, and prefers to keep most weight tensors at 2-bit whereas the activations range from 2-bit to 16-bit. 

\subsection{Experiments on Imagenet}
We ran an ablation study on ResNet18 \cite{heresidual} which is common in the quantization literature~\cite{jacob2018quantization,louizos2018relaxed,choi2018pact}. We started from the pretrained PyTorch model \cite{pytorch}.
We fine-tuned the model's weights jointly with the quantization parameters for 30 epochs using Bayesian Bits. During the last epochs  of Bayesian Bits training, BOP count remains stable but validation scores fluctuate due to the stochastic gates, so we fixed the gates and fine-tuned the weights and quantization ranges for another 10 epochs. 
To explore generalization to different architectures we experimented with the MobileNetV2 architecture \cite{mobilenetv2}, an architecture that is challenging to quantize \cite{nagel2019dfq, nagel2020adaround}.
The Appendix contains full experimental details, additional results, and a comparison of the results before and after fine-tuning.

\begin{figure*}[t!]
\centering
\begin{subfigure}[b]{0.49\linewidth}
  \centering
  \includegraphics[width=\linewidth]{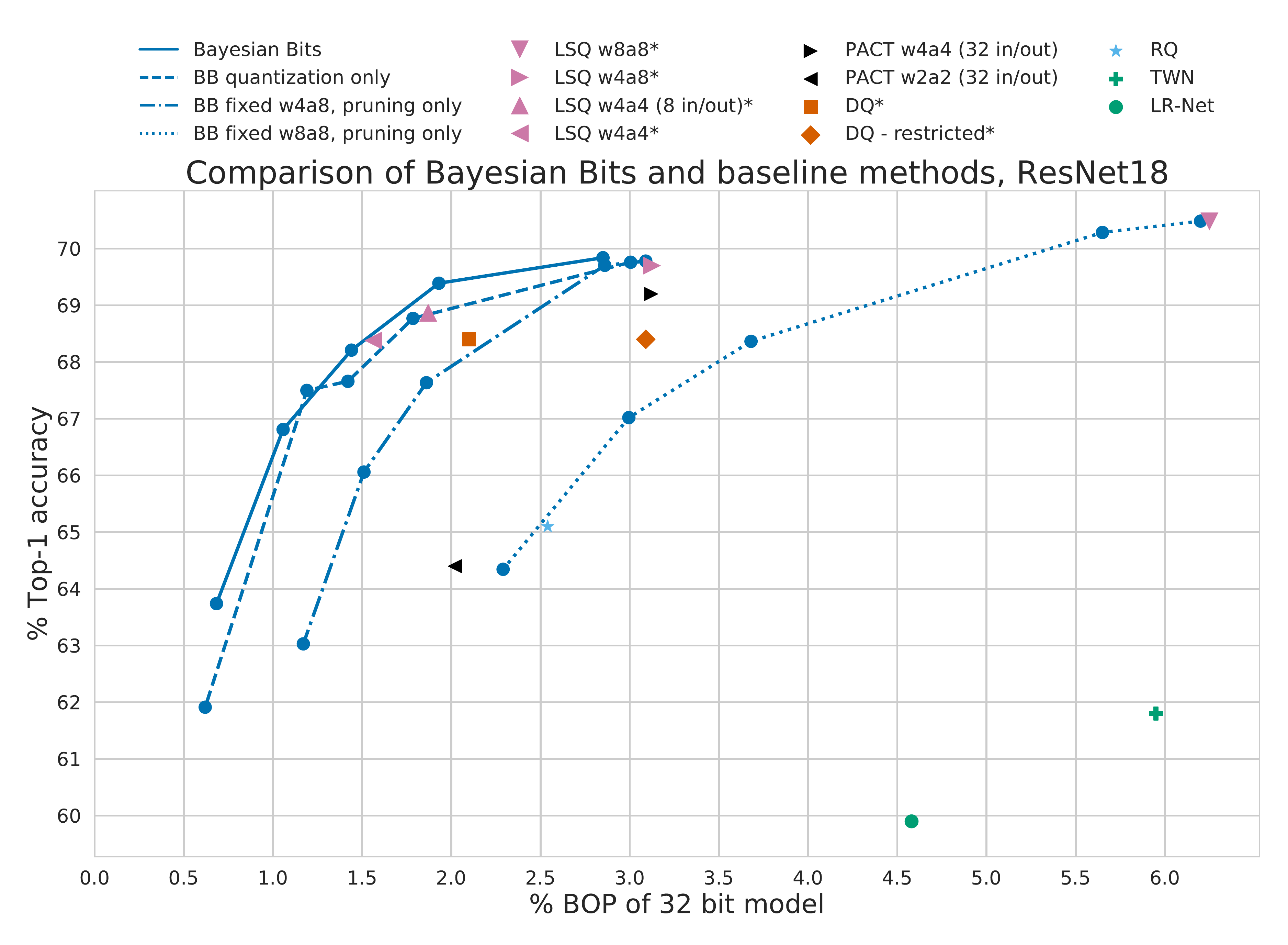}
  \caption{ResNet18 Imagenet results}\label{fig:bbplot}
 \end{subfigure}%
 ~
 \begin{subfigure}[b]{0.49\linewidth}
 \centering
  \includegraphics[width=\linewidth]{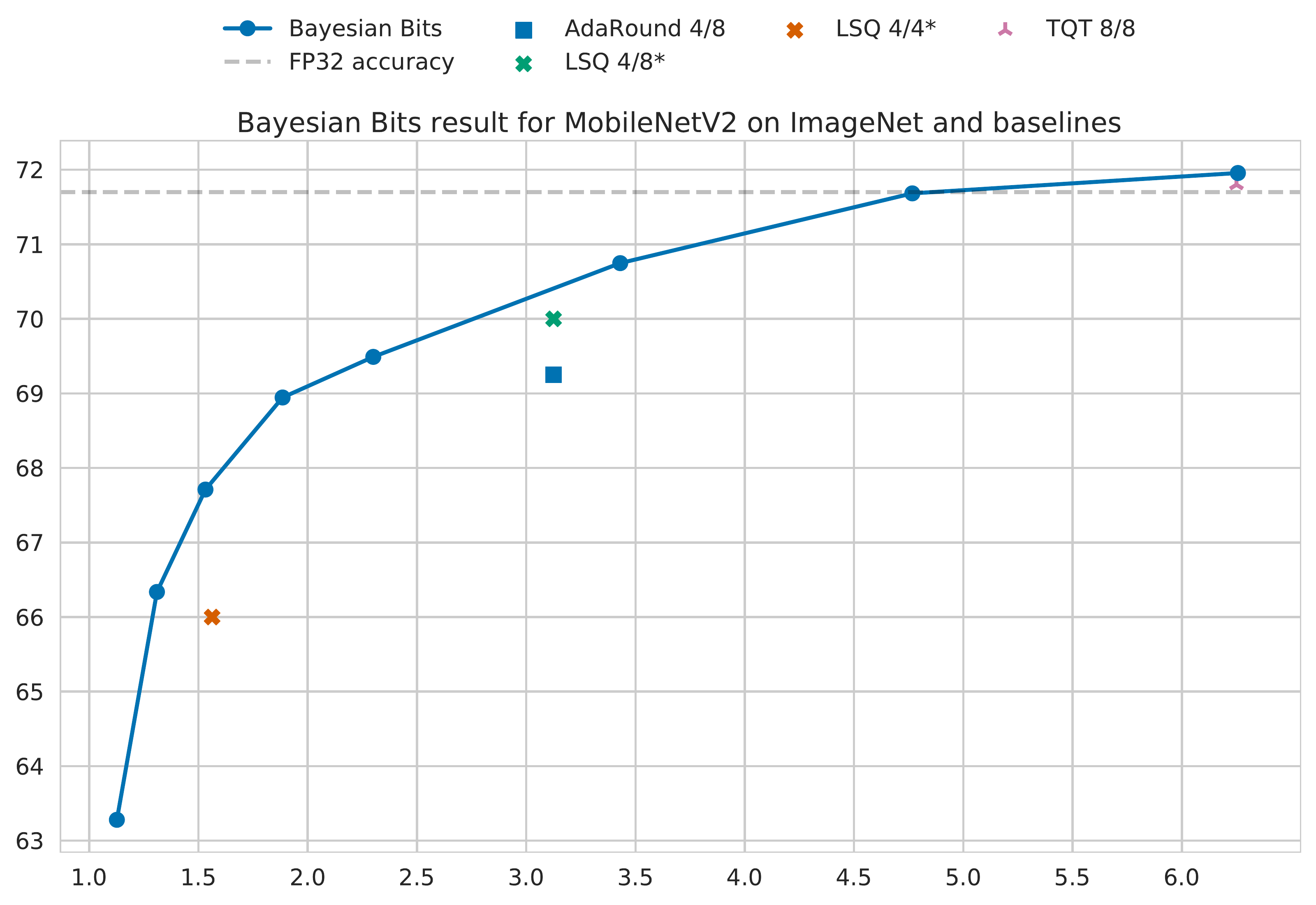}
  \caption{MobileNet V2 ImageNet results}\label{fig:bb-mnv2}
 \end{subfigure}
 \caption{\textbf{Imagenet Results.} (a) Bayesian Bits Imagenet validation accuracy on ResNet18. Bayesian Bits an BB Quantization only use $\mu\in\{0.03, 0.05, 0.07, 0.2\}$. BB pruning only uses $\mu \in\{0.05, 0.2, 0.5, 0.7, 1\}$ The Bayesian Bits results show the mean over 3 training runs. The quantization only and pruning only results show the mean over 2 training results. The BOP count per model is presented in the Appendix. The notation `wXaY' indicates a fixed bit width architecture with X bit weights and Y bit activations. `Z in/out' indicates that the weights of the first layer as well as the inputs and weights of the last layer are kept in Z bits. In this plot we additionally compare to PACT \citep{choi2018pact}. Note that PACT uses 32 bit input and output layers, which negatively affects their BOP count. In the Appendix we compare against a hypothetical setting in which PACT with 8 bit input and output layers yields the same results. * results run by the authors. (b) Bayesian Bits results on MobileNet V2, compared to AdaRound \citep{nagel2020adaround}, LSQ \citep{esser2019lsq}, and TQT \citep{jain2019tuq}  * results run by the authors.}
\end{figure*}

In Figure \ref{fig:bbplot} we compare Bayesian Bits against a number of strong baselines and show better trade offs between accuracy and complexity. 
We find different trade-offs by varying the global regularization parameter $\mu$.
Due to differences in experimental setup, we ran our own baseline experiments to obtain results for LSQ \cite{esser2019lsq}. 
Full details of the differences between the published experiments and ours, as well as experimental setup for baseline experiments can be found in the Appendix.

Besides experiments with combined pruning and quantization, we ran two sets of ablation experiments in which Bayesian Bits was used for pruning a fixed bit width model, and for mixed precision quantization only, without pruning. This was achieved through learning only the 4 bit and higher gates for the quantization only experiment, and only the zero bit gates for the pruning only experiment.
In \ref{fig:bbplot} we see that combining pruning with quantization yields superior results.

We provide the tables of the results in the Appendix along with visualizations of the learned architectures. Overall, we observe that Bayesian Bits provides better trade-offs between accuracy and efficiency compared to the baselines.
NB: we cannot directly compare our results to those of \cite{wang2019haq}, \cite{dong2019hawq,dong2019hawqv2} and \cite{wu2018mixed}, for reasons outlined in the Appendix, and therefore omit these results in this paper.

\vspace{-0.35cm}

\paragraph{A note on computational cost}
Bayesian Bits requires the computation of several residual error tensors for each weight and activation tensor in a network.
While the computational overhead of these operations is very small compared to the computational overhead of the convolutions and matrix multiplications in a network, we effectively need to store $N$ copies of the model for each forward pass, for $N$ possible quantization levels.
To alleviate the resulting memory pressure and allow training with reasonable batch sizes, we use gradient checkpointing \cite{chen2016checkpointing}.
Gradient checkpointing itself incurs extra computational overhead.
The resulting total runtime for one ResNet18 experiment, consisting of 30 epochs of training with Bayesian Bits and 10 epochs of fixed-gate fine-tuning, is approximately 70 hours on a single Nvidia TeslaV100.
This is a slowdown of approximately 2X compared to 40 epochs of quantization aware training.

\subsubsection{Post-training mixed precision}
\begin{figure}
    \centering
    \includegraphics[width=0.49\linewidth]{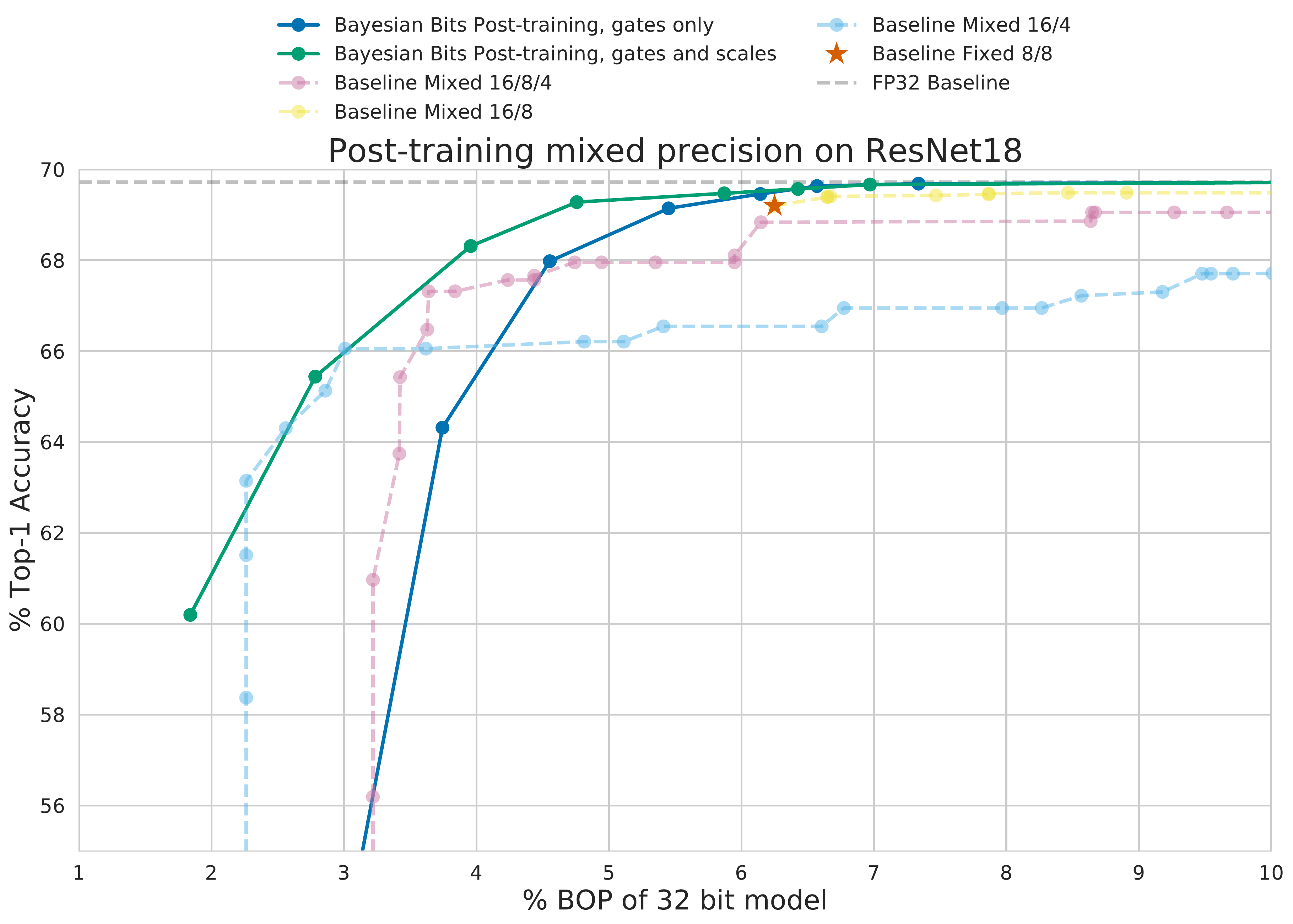}
    \caption{Pareto fronts of Bayesian Bits post-training and the baseline method, as well as a fixed 8/8 baseline}\label{fig:pareto}
\end{figure}
In this experiment, we evaluate the ability of our method to find sensible mixed precision settings by running two sets of experiments on a pre-trained ResNet18 model and a small version of ImageNet.
In the first experiment only the values of the gates are learned, while in the second experiment both the values of the gates and the quantization ranges are learned. 
In both experiments the weights are not updated.
We compare this method to an iterative baseline, in which weights and activation tensors are cumulatively quantized based on their sensitivity to quantization.
We compare against this baseline since it works similarly to Bayesian Bits, and note that this approach could be combined with other post-training methods such as Adaptive Rounding \cite{nagel2020adaround} after a global bit width setting is found.
Full experimental details can be found in the Appendix.
Figure~\ref{fig:pareto} compares the Pareto front of post-training Bayesian Bits with that of the baseline method and an 8/8 fixed bit width baseline \cite{nagel2019dfq}.
These results show that Bayesian Bits can serve as a method in-between `push-button' post-training methods that do not require backpropagation, such as~\cite{nagel2019dfq}, and methods in which the full model is fine-tuned, due to the relatively minor data and compute requirements.

\vspace{-0.2cm}
\section{Conclusion}\label{sec:conclusion}
\vspace{-0.2cm}
In this work we introduced Bayesian Bits, a practical method that can effectively learn appropriate bit widths for efficient neural networks in an end-to-end fashion through gradient descent. 
It is realized via a novel decomposition of the quantization operation that sequentially considers additional bits via a gated addition of quantized residuals. 
We show how to optimize said gates while incorporating principled regularizers through the lens of sparsifying priors for Bayesian inference. 
We further show that such an approach provides a unifying view of pruning and quantization and is hardware friendly. 
Experimentally, we demonstrated that our approach finds more efficient networks than prior art.

\medskip

\section*{Broader Impact}
Bayesian Bits allows networks to run more efficiently during inference time. This technique could be applied to any network, regardless of the purpose of the network. 

A positive aspect of our method is that, by choosing appropriate priors, a reduction in inference time energy consumption can be achieved. This yields longer battery life on mobile devices and lower overall power consumption for models deployed in production on servers.

A negative aspect is that quantization and compression could alter the behavior of the network in subtle, unpredictable ways.
For example, \cite{paganini2020fairness} notes that pruning a neural network may not affect aggregate statistics, but can have different effects on different classes, thus potentially creating unfair models as a result.
We have not investigated the results of our method on the fairness of the predictions of a model.

\begin{ack}
This work was funded by Qualcomm Technologies, Inc.
\end{ack}

\bibliography{bibliography}
\newpage

\title{Supplementary material of \\Bayesian Bits: Unifying Quantization and Pruning}
\author{}\date{}
\maketitle
\vspace{-2.3cm}
\appendix
\section{Further remarks on Bayesian Bits}
\subsection{Relationship to $L_0$ norm regularization}
It is interesting to see that the overall objective that we arrive at is similar to the stochastic version of the $L_0$ regularizer of~\citep{louizos2018learning}. By using the fact that
\begin{align}
    \sum_{i \in B}\prod_{j \in B}^{j \leq i} q_\phi(z_{jk} = 1) = \mathbb{E}_{q_\phi(\mathbf{z}_k)}\left[\sum_{i \in B}\prod_{j \in B}^{j \leq i} \mathbb{I}[z_{jk} \neq 0]\right],
\end{align}
we can rewrite Eq.~\ref{eq:bb_map} as follows:
\begin{align}
    \mathcal{F}(\theta, \phi )& := \mathbb{E}_{q_\phi(\mathbf{z}_{1:K})}\Bigg[\frac{1}{N}\log p_\theta(\mathcal{D}|\mathbf{z}_{1:K}) - \nonumber \lambda^\prime \sum_k \sum_{i \in B}\prod_{j \in B}^{j \leq i} \mathbb{I}[z_{jk} \neq 0]\Bigg]. \label{eq:bb_map_withind}
\end{align}
Now by assuming that the parameters will not be quantized, i.e. $z_4 = z_8 = z_{16} = z_{32} = 1$ the objective becomes 
\begin{align}
    \mathbb{E}_{q_\phi(\mathbf{z}_{2,1:K})}\left[\frac{1}{N}\log p_\theta(\mathcal{D}|\mathbf{z}_{2,1:K}) - \lambda^\prime |B| \sum_k \mathbb{I}[z_{2k} \neq 0]\right],
\end{align}
which corresponds to regularizing with a specific strength the expected $L_0$ norm of the vector that determines the group of parameters that will be included in the model.

\subsection{Optimizing the Bayesian bits objective}
For optimization, we can exploit the alternative formulation of the Bayesian bits objective, presented at Eq.~\ref{eq:bb_map_withind}, to use the hard-concrete relaxation of~\cite{louizos2018learning}. More specifically, the hard concrete distribution has the following sampling process:
\begin{align}
    u_{jk} \sim & U[0, 1], \hspace{0.25em} g_{jk} = \log \frac{u_{jk}}{1 - u_{jk}}, \hspace{0.25em} s_{jk} = \sigma\left(\frac{g_{jk} + \phi_{jk}}{\tau}\right)\nonumber\\
    & z_{jk} = \min(1, \max(0, s_{jk} (\zeta - \gamma) + \gamma))\label{eq:sample_hc}
\end{align}
where $\sigma(\cdot)$ corresponds to the sigmoid function, $\tau$ is a temperature hyperparameter and $\zeta, \gamma$ are hyperaparameters that ensure $z$ has support for exact $0, 1$. Essentially, it corresponds to a mixture distribution that has three components: one that corresponds to zero, one that corresponds to one and one that produces values in $(0, 1)$. Under this relaxation, the objective in Eq.~\ref{eq:bb_map_withind} will be converted to
\begin{align}
    \mathcal{F}(\theta, \phi )& := \mathbb{E}_{r_\phi(\mathbf{z}_{1:K})}\left[\frac{1}{N}\log p_\theta(\mathcal{D}|\mathbf{z}_{1:K})\right] - \nonumber \lambda^\prime \sum_k \sum_{i \in B}\prod_{j \in B}^{j \leq i} R_\phi(z_{jk} > 0).\label{eq:bb_relaxed}
\end{align}
where $R_\phi(\cdot)$ corresponds to complementary cumulative distribution function, i.e. $1 - R_\phi(\cdot)$ is the cumulative distribution function (CDF), of the density $r_\phi(z)$ induced by the sampling process described at Eq.~\ref{eq:sample_hc}. The $R_\phi(z_{jk} > 0)$ now corresponds to the probability of activating the gate $z_{jk}$ and has the following simple form 
\begin{align}
    R_\phi(z_i > 0) = \sigma\left(\phi - \tau \log\frac{-\gamma}{\zeta}\right).
\end{align}
For the configuration of the gates at test time we use the following expression which results into $z \in \{0, 1\}$
\begin{align}
    z = \mathbb{I}\left[\sigma\left(\tau \log\left(-\frac{\gamma}{\zeta}\right) -\phi\right) < t\right],
\end{align}\label{eq:thresholding}
where we set $t = 0.34$. This threshold value corresponds to the case when the probability of the mixture component corresponding to exact zero is higher than the other two.

\subsection{Alternative gating approaches}
\begin{table}[h]
\small
\centering
\caption{Results of experiments on Bayesian Bits with deterministic gates. For the first phase of training we lowered the learning rate of the gate parameters to $10^{-4}$, and initialized the gate parameters to $2$, which is much closer to the saturation point of the HardSigmoid function, and kept the other hyperparameters the same. Pre-FT results of the deterministic gate experiments and post-FT Bayesian Bits results are included for comparison. }
\begin{tabular}{llccc}
\toprule
Experiment & Gating type & Acc. (\%) & Rel. GBOPs (\%) & CE Loss \\ 
\midrule
CIFAR10, VGG, $\mu=0.01$ & Stochastic & 93.23$\pm$0.10 & 0.51$\pm$0.03 & 0.00$\pm$0.00 \\
CIFAR10, VGG, $\mu=0.01$ & Deterministic & 92.82 & 0.42 & 0.00 \\
\midrule
ImageNet, ResNet18, $\mu=0.03$ & Stochastic & 69.36$\pm$0.11 & 1.93$\pm$0.05 & 1.26$\pm$0.00 \\
ImageNet, ResNet18, $\mu=0.2$ & Stochastic & 63.76$\pm$0.36 & 0.68$\pm$0.03& 1.64$\pm$0.03 \\
ImageNet, ResNet18, $\mu=0.03$ (Pre-FT) & Deterministic& 0.24 & 0.30 & 1.88 \\
ImageNet, ResNet18, $\mu=0.03$& Deterministic & 56.81 & 0.30 & 1.88 \\
ImageNet, ResNet18, $\mu=0.03$ (Pre-FT) & Deterministic & 6.74 & 24.09 & 1.49 \\
ImageNet, ResNet18, $\mu=0.03$ & Deterministic & 68.03 & 24.09 & 1.49 \\
ImageNet, ResNet18, $\mu=0.03$ (Pre-FT) & Deterministic & 54.61 & 0.48 & 1.96 \\
ImageNet, ResNet18, $\mu=0.03$ & Deterministic & 60.18 & 0.48 & 1.96 \\

\bottomrule
\end{tabular}
\label{tab:deterministic}
\end{table}

We experimented with several other gating approaches.
In this section we describe these approaches, their downsides when compared to the approach described in section 2, and results to support our claims where possible.

\paragraph{Deterministic gates} We experimented with deterministic gates. Results of preliminary experiments with deterministic gates can be found in Table \ref{tab:deterministic}. Here we can see that while deterministic gates do not significantly hurt results on the CIFAR 10 experiments, ImageNet experiments do not fare so well. We attribute this to the following observations: 1) Gates getting ``stuck'': once a deterministic gate parameter assumes a value in the saturated 0 part of the hardsigmoid function, it no longer receives any gradients from the cross-entropy loss.  Due to the stochasticity in our gates there is always a nonzero probability that a gate will be (partially) on and receive a gradient from the loss. 2) The model can learn to keep deterministic gates fixed at a value between 0 and 1 during training, and essentially use it as a free parameter to reduce the cross-entropy loss, while simulatenously lowering the regularization loss. This creates a disconnect between the training and the inference models, as during inference we fix the gates to either 0 or 1, as described in Section 2.  This effect can be seen in Table \ref{tab:deterministic}: for the ResNet experiments we see (pre fine-tuning) training loss values usually associated with much higher validation accuracies. The model can compensate for this through additional fine-tuning, but we note that for the deterministic gate experiments the same hyperparameters gave strongly differing results, which we did not observe for the stochastic gates. 
We experienced the same issues for deterministic non-saturating sigmoid gates.

\paragraph{REINFORCE} We experimented with vanilla REINFORCE, and REINFORCE enhanced with several standard baselines, but found that the high variance of the estimated gradients posed difficulties for optimization of our networks, and abandoned this route.

\subsection{Bayesian Bits algorithm}
At Figure~\ref{fig:alg_bb} we provide the algorithm for the forward pass with a Bayesian Bits quantizer.
\begin{figure*}[ht!]
	\null\hfill 
	\begin{minipage}[t]{.5\linewidth}
	    \begin{algorithm}[H]
	        \caption{Forward pass with Bayesian bits} 
            \label{alg:bayesian_bits_forward}
			\begin{algorithmic}
			    \REQUIRE Input $x$, $\alpha, \beta$, $\phi$
				\STATE clip(x, $\min=\alpha$, $\max = \beta$)\\
				\STATE $s_2 \gets \frac{\beta - \alpha}{2^2 - 1}$, \enskip $x_2 \gets s_2 \lfloor \frac{x}{s_2}\rceil$\\
				\STATE $z_2 \gets $ get\_gate($\phi_2$), \enskip $x_q \gets z_2 x_2 $
				\STATE \FOR {$b$ in $\{4, 8, 16, 32\}$}
				\STATE $s_b \gets \frac{s_{b/2}}{2^{b/2} + 1}$, \enskip $z_b \gets $ get\_gate($\phi_b$)\\
				\STATE $\epsilon_b \gets s_b \bigg\lfloor \frac{x - \left(x_2 + \sum_{j  < b}\epsilon_j\right)}{s_b} \bigg\rceil$\\
				\STATE $x_q \gets x_q + z_b \left(\prod_{j < b}z_j\right)\epsilon_b$
				\ENDFOR\\
				\STATE \textbf{return } $x_q$
			\end{algorithmic}%
		\end{algorithm}%

	\end{minipage}%
	\hfill 
	\begin{minipage}[t]{.5\linewidth}
		\begin{algorithm}[H]
			\caption{Getting the gate during training and inference}
			\label{alg:get_gate}
			\begin{algorithmic}
				\REQUIRE Input $\phi, \zeta, \gamma, \beta, t$, training \\
				\IF{training}
				\STATE $u \sim U[0, 1]$, \enskip $g \gets \log \frac{u}{1 - u}$, \enskip $s \gets \sigma((g + \phi) / b)$\\
				\STATE $z \gets \min(1, \max(0, s(\zeta - \gamma) + \gamma))$
				\ELSE
				\STATE $z \gets \mathbb{I}\left[\sigma\left(\beta \log\left(-\frac{\gamma}{\zeta}\right) -\phi\right) < t\right]$ 
				\ENDIF
				\STATE \textbf{return } $z$
			\end{algorithmic}%
		\end{algorithm}%
	\end{minipage}
	\caption{Pseudo-code for the forward pass of the Bayesian Bits quantizer.}\label{fig:alg_bb}
\end{figure*}

\subsection{Decomposed quantization for non-doubling bit widths}
Consider the general case of moving from bit width $a$ to bit width $b$, where $0<a<b$, for a given range $[\alpha, \beta]$.
Using the equation of section 2.1, i.e. $s_b=s_a / {2^{b-a}-1}$ yields a value of $(\beta-\alpha)/N$, where $N=2^b+2^a-2^{b-a}-1$. 
If $b=2a$ then this simplifies to $N=2^b-1$, which is the desired result.
However, if $b\neq 2a$ then there are two cases to distinguish:
\begin{enumerate}
    \item $b > 2a$, in this case we can write $N=2^{2a+c}+2^a-2^{a+c}-1$, where $c=b-2a$.
    There are $2^{a+c}-2^a$ bins more than desired in the range.
    \item $b < 2a$, in this case we can write $N=2^{2a-c}+2^a-2^{a-c}-1$, where $c=2a-b$.
    There are $2^a-2^{a-c}$ fewer bins than desired.
\end{enumerate}

In these cases $\alpha$ and $\beta$ must be scaled according to the difference between the expected and the true number of bins.

\section{Experimental details}

\subsection{Experimental setup}\label{sec:expsetup}
The LeNet-5 model is realized as 32C5 - MP2 - 64C5 - MP2 - 512FC - Softmax, whereas the VGG is realized as 2x(128C3) - MP2 - 2x(256C3) - MP2 - 2x(512C3) - MP2 - 1024FC - Softmax. The notations is as follows: 128C3 corresponds to a convolutional layer of 128 feature maps with 3x3 kernels, MP2 corresponds to max-pooling with 2x2 kernels and a stride of 2, 1024FC corresponds to a fully connected layer with 1024 hidden units and Softmax corresponds to the classifier. Both models used ReLU nonlinearities, whereas for the VGG we also employed Batch-normalization for every layer except the last one. The weights, biases, gates, and ranges were optimized with Adam~\citep{kingma2014adam} using the default hyper-parameters for $100$ epochs with a batch size of 128 on MNIST, $300$ epochs with a batch size of 128 on CIFAR 10 and during the last $1/3$ epochs we linearly decayed the learning rate to zero. For CIFAR 10, we also performed standard data augmentation: random horizontal flips, random crops of 4 pixel padded images, and channel standardization. For the test images, we only performed channel standardization. 
We do not perform additional fine-tuning with fixed gates for the MNIST and CIFAR 10 epxeriments, as we found this did not improve results.

For the ResNet18 we used SGD with a learning rate of 3e-3 and Nesterov momentum of 0.9 for the network parameters and used Adam with the default hyperparameters for the optimization of the gate parameters and ranges. The learning rates for all of the optimizers were decayed by a factor of 10 after every 10 epochs. We did not employ any weight decay and used a batch-size of 384 distributed across four Tesla V100 GPUs. 
After training we fixed the gates using the thresholding described in Eq \ref{eq:thresholding}, and fine-tuned the weights and scale parameter $\beta$ for 10 epochs. 
In this stage we used SGD Nesterov momentum of 0.9 for the weights, and Adam for the scales, both starting at a learning rate of $10^{-4}$ and annealed to 0 using cosine learning rate annealing at each iteration.

\subsection{BOP and MAC count}
The BOP count of a layer $l$ is computed as:
\begin{align}
    \text{BOPs}(l) = \text{MACs}(l)b_wb_a, \label{eq:comp_bop_count_app}
\end{align}\label{eq:bopcount}
where $b_w$ is the bit width of the weights and $b_a$ is the bit width of the (input) activations.

\subsubsection{BOP-aware regularization}
We set the regularization strength for each gate $z_{jk}$ to be $\mu \lambda^\prime_{jk}$, where $\lambda^\prime_{jk}$ is proportional to the BOP count corresponding to the bit width $j$ and the MAC count of the layer $l_k$ that the quantizer $k$ operates on. Specifically, we set $\lambda^\prime_{jk} = b_j \text{ MACs}(l_k) / \max([\text{MACs}(1), \dots, \text{MACs}(L)])$, where $b_j$ is the bit width that gate $j$ controls and $L$ corresponds to the total number of layers. 

In practical applications, one would experiment with a range of regularization strengths to generate a Pareto curve, and pick a model that achieves a suitable tradeoff between target task performance and BOP.
We leave targeting a specific BOP count for future work.

\subsubsection{BOP and MAC count under sparsity}
Since the sparsification only affects a layer's MAC count and not its bit width, Eq \ref{eq:bopcount} holds for sparsified networks as well. 
However, it is insightful to see how sparsity affects a layer's BOP count through its effect on the layer's MAC count.

The MAC count of a convolutional layer can be derived as follows.
For each output pixel in a feature map we know that $W_f\times W_h\times B$ computations were performed, where $W_f$ and $W_h$ are the filter width and height, and $B$ is the convulational block size (e.g. for dense convolutions $B$ is equal to the number of input channels, for depthwise separable convolutions $B$ is equal to 1).
There are $C_o \times W \times H$ output pixels, where $C_o$ is the number of output channels, and $W$ and $H$ are the width and height of the output map. 
Henceforth we only consider dense convolutional layers, i.e. layers where $B=C_{i}$ where $C_{i}$ is the number of input channels.
Thus, the MAC count of a convolutional layer $l$ can be computed as $\text{MACs}(l)=C_o \times W \times H \times C_i \times W_f \times H_f$.
Note that in this formulation, no special care needs to be taken in considering padding, stride, or dilations. 

As stated earlier, pruning output channels of layer $l-1$ corresponds to pruning the associated activations, which in turn corresponds to pruning input channels of layer $l$. 
If we assume that $C_{i'}$ output channels are maintained in layer $l-1$, and $C_{o'}$ output channels are maintained in layer $l$, the pruned MAC count can be computed as:

\begin{align}
    \text{MACs}_\text{pruned}(l) &= p_i C_i p_o C_o W H W_f H_f \\
    &= p_i  p_o \text{MACs}(l)
\end{align}

where $p_i = C_{i'}/C_i, p_o=C_{o'}/C_o$, and $\text{MACs}(l)$ is used to denote the MAC count of the unpruned layer.
As a result, if we know the input and output pruning ratios $p_i$ and $p_o$, the BOP count can be computed without recomputing the MAC count for the pruned layers with a slight modification of equation \ref{eq:comp_bop_count_app}:

\begin{align}
    \text{BOPs}_\text{pruned}(l) &= \text{MACs}_\text{pruned}(l)b_wb_a \\
        &=p_ip_o\text{MACs}(l)b_wb_a\label{eq:bop_pruned}
\end{align}

\subsubsection{ResNet18 MAC count computation}
To compute the BOP count for ResNet18 models, we need to be careful with our application of equation \ref{eq:bop_pruned}, due to the presence of residual connections: to turn off an input channel at the input of a residual block, it must be turned off both in the output of the previous block as well as in the residual connection.
We circumenvent this issue by only considering $p_i$ for the inputs of the second convolutional layer in each of the residual blocks, where there is no residual connection.
Elsewhere, $p_i$ is always assumed to be 1.
Output pruning is treated as in any other network, since the removal of output channels always leads to reduced MAC count.
Thus, the BOP counts reported for ResNet18 models must be interpreted as an upper bound; the real BOP count may be lower.

\subsubsection{ResNet18 regularization}
In ResNet architectures, the presence of downsample layers means that certain quantized activation tensors feed into two multiple convolution operations, i.e. the downsample layer and the input layer of the corresponding block.
As a result, we need to slightly modify the computation of $\lambda'_{jk}$ as introduced in Section 4 for these activation quantizers.
For an activation quantizer $k$ for which this is the case, we compute $\lambda'_{jk}$ as follows:

\begin{equation}
    \lambda'_{jk}=b_j \frac{\left(\text{MACs}(l_d) + \text{MACs}(l_c)\right)}{\max\left([\text{MACs}(1)\dots,\text{MACs}(L)]\right)}
\end{equation}

where $l_d$ and $l_c$ denote the downsample layer and the first convolutional layer in the corresponding block respectively.

\section{Baselines}

\subsection{Differences between baselines and our experimental setup}

\begin{table}[h]
    \small
    \centering
    \caption{Differences in experimental setup between the LSQ \cite{esser2019lsq} and PACT\cite{choi2018pact} baselines and ours}.
    \begin{tabular}{lllllc}
        \toprule
        Experiment & ResNet18 type & BN & Act quant & Grad scaling & FP32 acc \\ 
        \midrule
        LSQ \cite{esser2019lsq} & Pre & Not folded & Input & Yes & 70.5\% \\
        PACT \cite{choi2018pact} & Pre & ? & Input & No & 70.2\% \\
        \midrule
        LSQ (our impl) & Post & Folded \cite{krishnamoorthi2018quantizing} & Output & No & 69.7\% \\
        Bayesian Bits & Post & Folded \cite{krishnamoorthi2018quantizing} & Output & No & 69.7\% \\
        \bottomrule
    \end{tabular}
    \label{tab:baseline_diffs}
\end{table}

In Table \ref{tab:baseline_diffs} we show the differences in experimental setup between our experiments and the experimental setup as used in LSQ and PACT. 
In Table \ref{tab:baseline_diffs} we compare along the following axes: 
usage of pre- or post-activation ResNet18; pre-activation ResNet18 gives higher baseline accuracy. 
Handling of batch normalization layers; not folding BN parameters into the associated weight tensors is identical to using per-channel quantization, instead of per tensor quantization. 
Per-channel quantization gives higher accuracy than per-tensor quantization. 
Input or output quantization: Using input quantization implies that activation tensors are not quantized until they are used as input to an operation. This in turn implies that high-precision activation tensors need to be stored and thus transported between operations. 
This does not affect network BOP count but might yield increased latency for hardware deployment.
FP32 accuracy: higher FP accuracy on the same architecture is likely to yield higher quantized accuracy. 
Gradient scaling: this is a technique introduced by \cite{esser2019lsq}.

There are several works of note to which we cannot directly compare our results.
\cite{wang2019haq} only present ImageNet results on ResNet50 \cite{heresidual} and MobileNet \cite{mobilenetv2} architectures.
Furthermore, the authors do not provide BOP counts for their models, making direct comparison to our results impossible.
\cite{dong2019hawq,dong2019hawqv2} and \cite{wu2018mixed} do present Imagenet results on ResNet18, but do not provide the mixed precision configuration for their reported results. 
While \cite{wu2018mixed} provide the BOP count of the resulting ResNet18 network, it is not mentioned whether the fact that the first and last layers are in full precision is taken into account in determining the compute reduction. Furthermore, they include a 3-bit configuration in their search space, which is not efficiently implemented in hardware.
This makes it hard to compute the BOP count using Eq.~\ref{eq:bopcount}.
Furthermore, their models are optimized for weight size reduction, not  compute reduction.

\paragraph{LSQ Experimental Details}
A fair comparison between the published results of \cite{esser2019lsq} and our results is not possible due to the differences in experimental setup highlighted in Table~\ref{tab:baseline_diffs}. 
To ensure that we would still do the baseline method of \cite{esser2019lsq} justice, we ran an extensive suite of experiments to optimize the experimental hyperparameters.
The results presented in Figure~\ref{fig:bbplot} and Table~\ref{tab:res_imagenet} are obtained as follows: we trained the network parameters and scales with Adam optimizers, with the same learning rate for the parameters and the scales.
We performed grid search over the learning rate.
The best learning rates were $10^{-3}$ for the w8a8 experiment, $10^{-5}$ for w4a8, $10^{-4}$ for w4a4 with 8 bit inputs and outputs, and $3\cdot10^{-5}$ for full w4a4.
We experimented with using SGD for the network parameters while using Adam with a lower learning rate for the scales, but found that using Adam for both consistently yielded better performance.
We trained for 40 epochs, and decayed all learning rates using cosine decay to $10^{-3}$ times the original learning rate, except for the w8a8 experiment where we trained for 20 epochs and annealed to $10^{-2}$ times the original learning rate.
In experiments with 8 bit weights we initialized the weight quantization ranges using the minimum and maximum of the weight tensor.
For 8 bit activations we initialized the quantization ranges using an exponential moving average of the per-tensor minimum and maximum over a small number of batches.
For 4 bit quantization we performed grid search over ranges to find the range that minimizes the mean squared error between the FP32 and quantized values of the weight and activation tensors.
We applied a weight decay of $10^{-4}$ in all our experiments.1
We folded Batch normalization parameters into the preceding weight tensors prior to training.
We did not use gradient scaling as we did not use it in our own experiments.

\section{Further Results}

\subsection{Updated ImageNet results}
\begin{figure}[h]
  \centering
  \includegraphics[width=\linewidth]{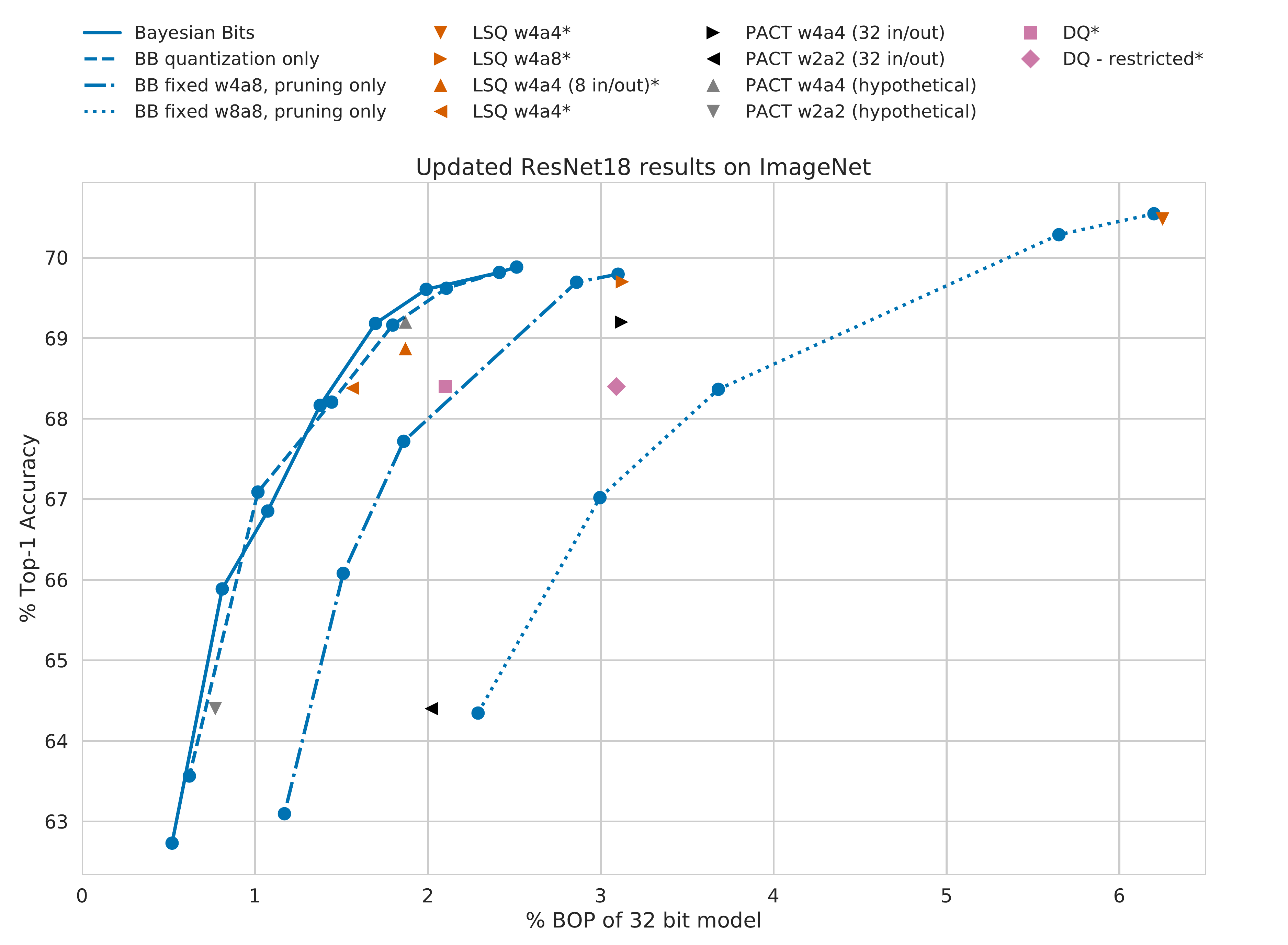}
  \caption{Bayesian Bits ResNet18 ImageNet results. In these experiments the activations feeding into a residual connection were not quantized, contrary to the results presented in \ref{fig:bbplot}.}\label{fig:imagenet_new_results}
\end{figure}

The ResNet18 model used in the experiments of Figure~\ref{fig:bbplot} and Table~\ref{tab:res_imagenet} quantized the activations that feed into residual connections. 
Since the bit-widths of these quantizers do not affect the BOP count, the quantizers were effectively over-regularized. 
To assess what the effect of over-regularization on these quantizers was, we ran a new set of experiments in which these activations were not quantized. 
The results of these experiments are plotted in \ref{fig:imagenet_new_results} and included in \ref{tab:res_imagenet} (experiments labeled Updated). 
This change only affects the Bayesian Bits and BB Quant only results.
Note that for this arguably more realistic scenario, our method more clearly outperforms the baselines.

\subsection{MNIST and CIFAR 10}
\begin{figure*}[ht!]
\centering
\begin{subfigure}[b]{0.32\linewidth}
  \centering
  \includegraphics[width=\linewidth]{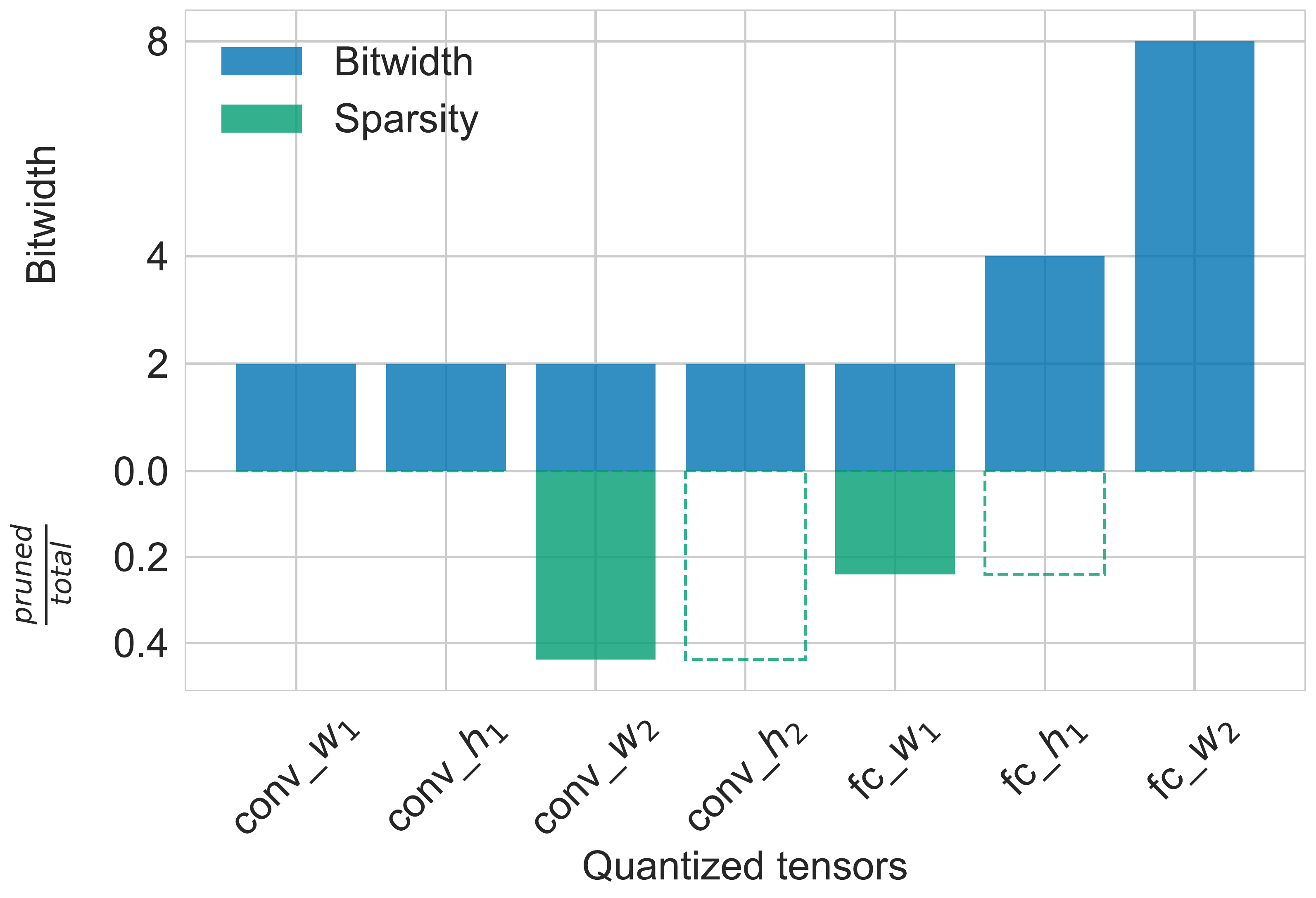}
  \caption{}
 \end{subfigure}%
 ~
 \begin{subfigure}[b]{0.32\linewidth}
  \centering
  \includegraphics[width=\linewidth]{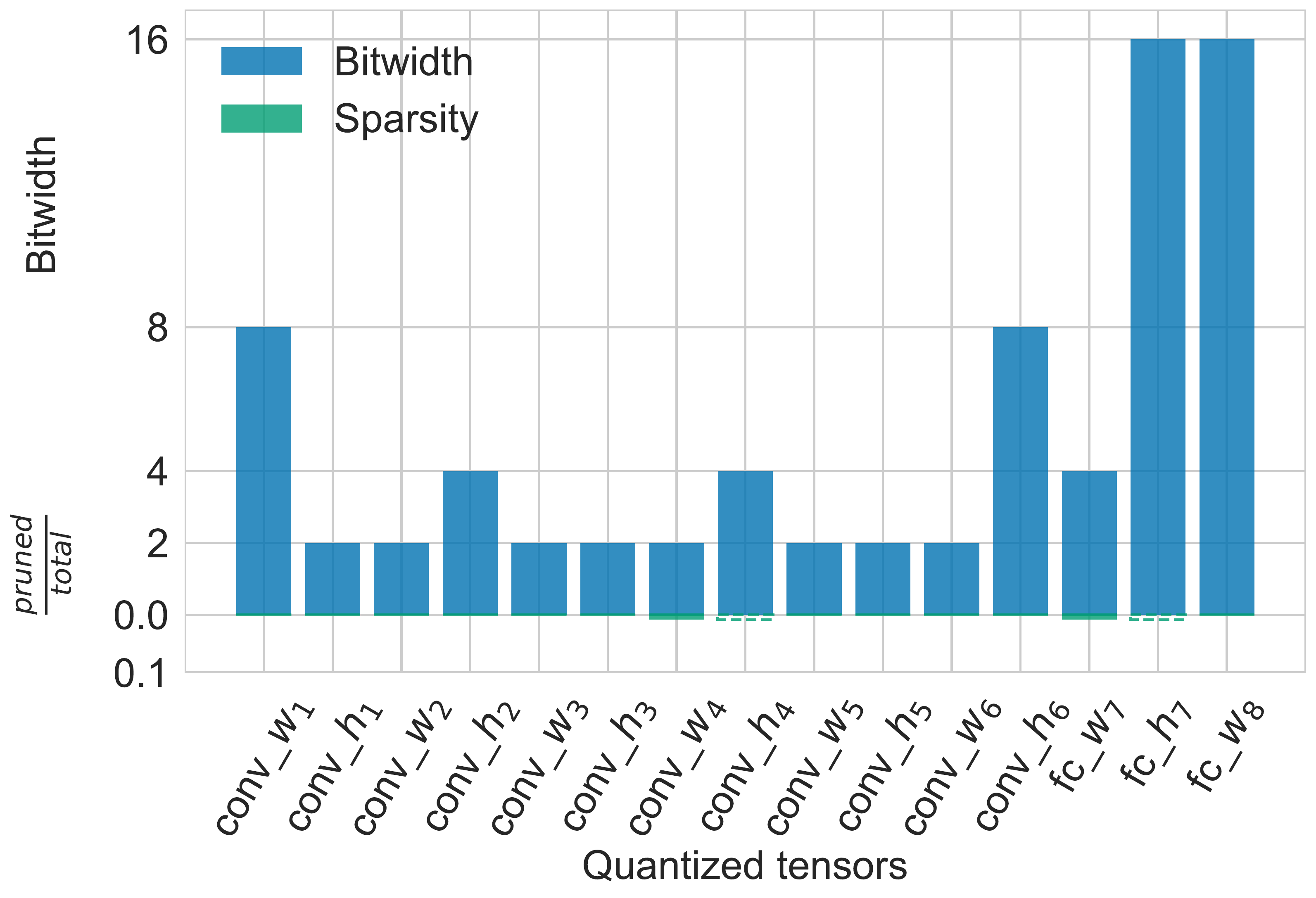}
  \caption{}
 \end{subfigure}%
 ~
 \begin{subfigure}[b]{0.32\linewidth}
 \centering
  \includegraphics[width=\linewidth]{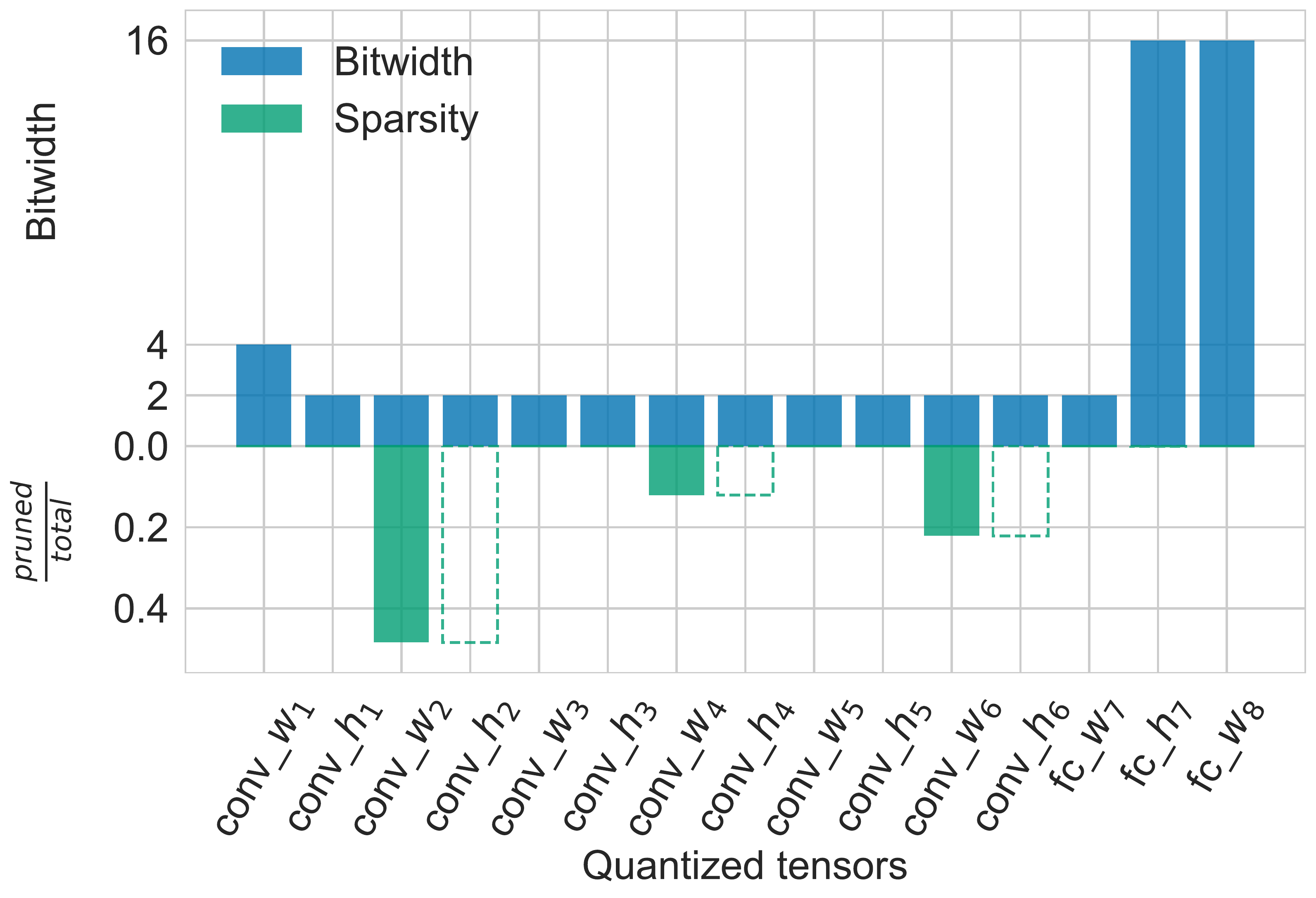}
  \caption{}
 \end{subfigure}
 \caption{\textbf{Learned LeNet-5 and VGG architectures.} (a) Illustrates the bit-allocation and sparsity levels for the LeNet-5 whereas (b) illustrates the bit-allocation and sparsity levels for the best performing VGG, accuracy wise. (c) Illustrates a VGG model trained with more aggressive regularization, resulting into less BOPs and more quantization / sparsity. With the dashed lines we show the implied sparsity on the activations due to the sparsity in the (preceding) weight tensors.}\vspace{-0.30cm}
 \label{fig:learn_arch_toy}
\end{figure*}

Figure~\ref{fig:learn_arch_toy} shows the learned bit width and sparsity per quantizer. Note that structural sparsity is only applied to weight quantizers, which implicitly applies it to activation tensors as well.

\subsection{Effect of fine-tuning}
\begin{figure}[h]
  \centering
  \includegraphics[width=\linewidth]{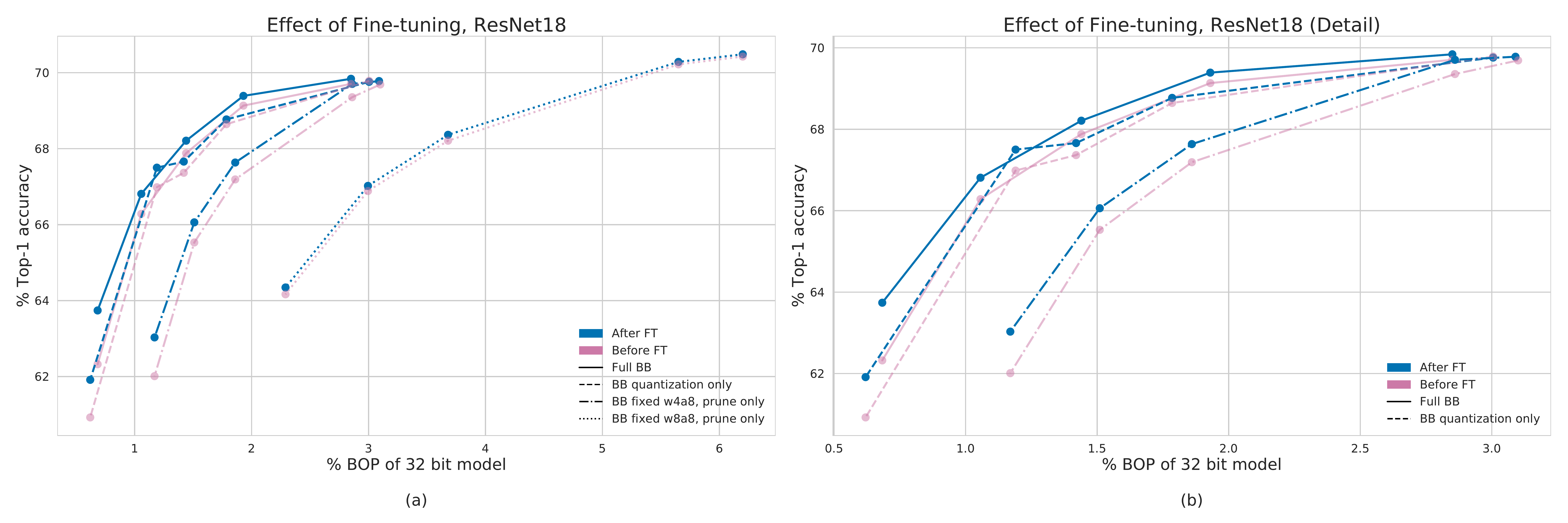}
  \caption{Bayesian Bits Imagenet validation accuracy on ResNet18 before and after final 10 epochs of fine-tuning for $\mu\in\{0.03, 0.05, 0.07, 0.2\}$. Means and individual runs of 3 training runs for each $\mu$. Plot (b) contains a close-up of results of the full Bayesian Bits, quantization only, w4a8 prune only experiments. }\label{fig:imagenet_pre_post_ft}
\end{figure}

The effects of fine-tuning on final model accuracy are presented in Figure~\ref{fig:imagenet_pre_post_ft} and in Table~\ref{tab:res_imagenet}.

\subsection{ImageNet}
\begin{figure}[h]
  \centering
  \includegraphics[width=0.9\linewidth]{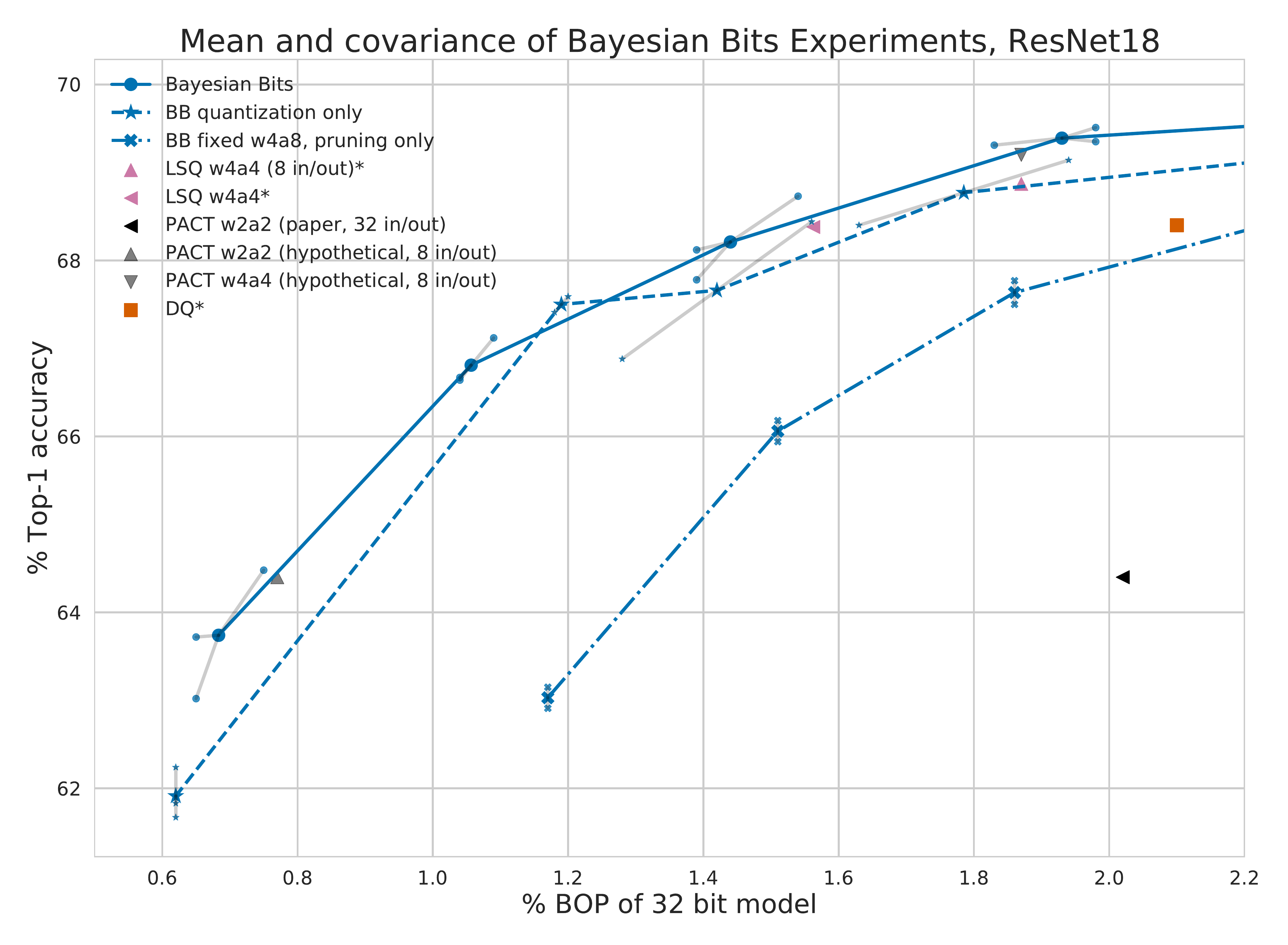}
  \caption{Bayesian Bits Imagenet validation accuracy on ResNet18 for $\mu\in\{0.03, 0.05, 0.07, 0.2\}$. Means and individual runs of 3 training runs for each $\mu$. The PACT \cite{choi2018pact} marked with 'hypothetical' are hypothetical results, in which the BOP count was computed using 8 bit input and output layers, instead of the full precision input and output layers used in \cite{choi2018pact}, and we make the optimistic assumption that this would not produce different results. }\label{fig:imagenet_scratch_results}
\end{figure}

\begin{figure*}[ht!]
  \centering
  \includegraphics[width=\linewidth]{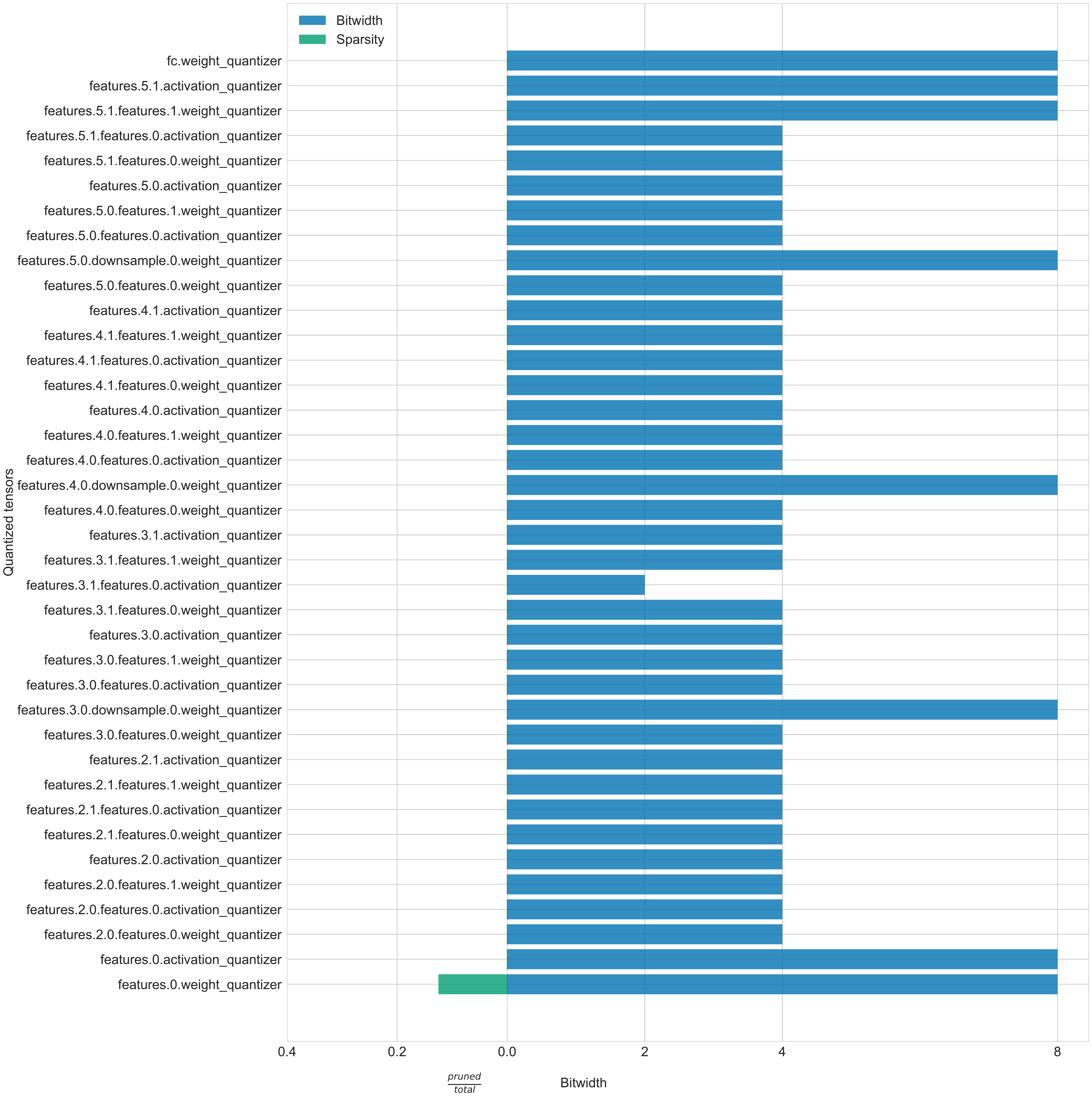}
  \caption{Learned ResNet18 architecture for first run with $\mu = 0.03$.}\label{fig:resnet_003}
\end{figure*}

\begin{figure*}[ht!]
  \centering
  \includegraphics[width=\linewidth]{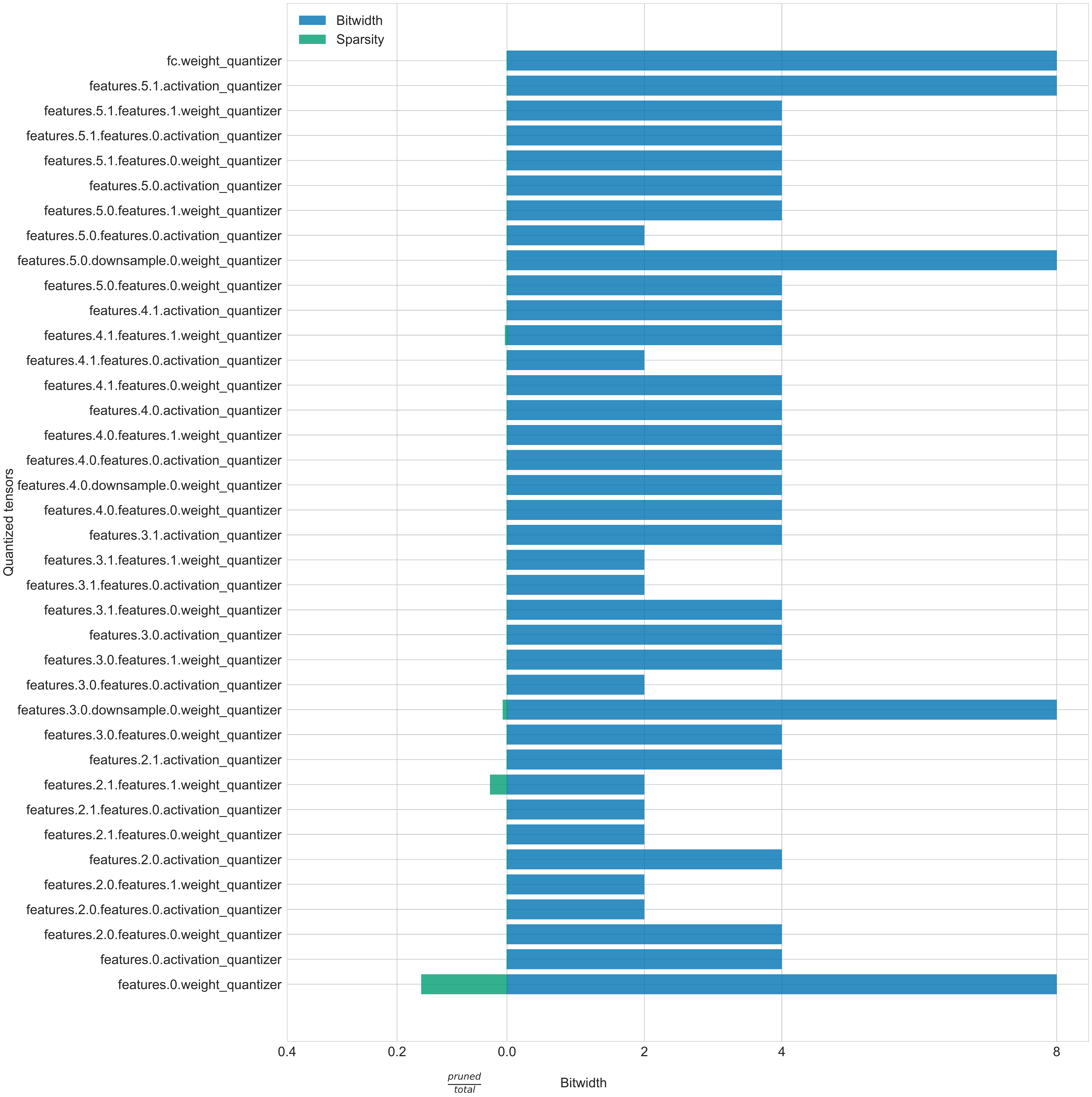}
  \caption{Learned ResNet18 architecture for first run with $\mu = 0.05$.}\label{fig:resnet_005}
\end{figure*}

\begin{figure*}[ht!]
  \centering
  \includegraphics[width=\linewidth]{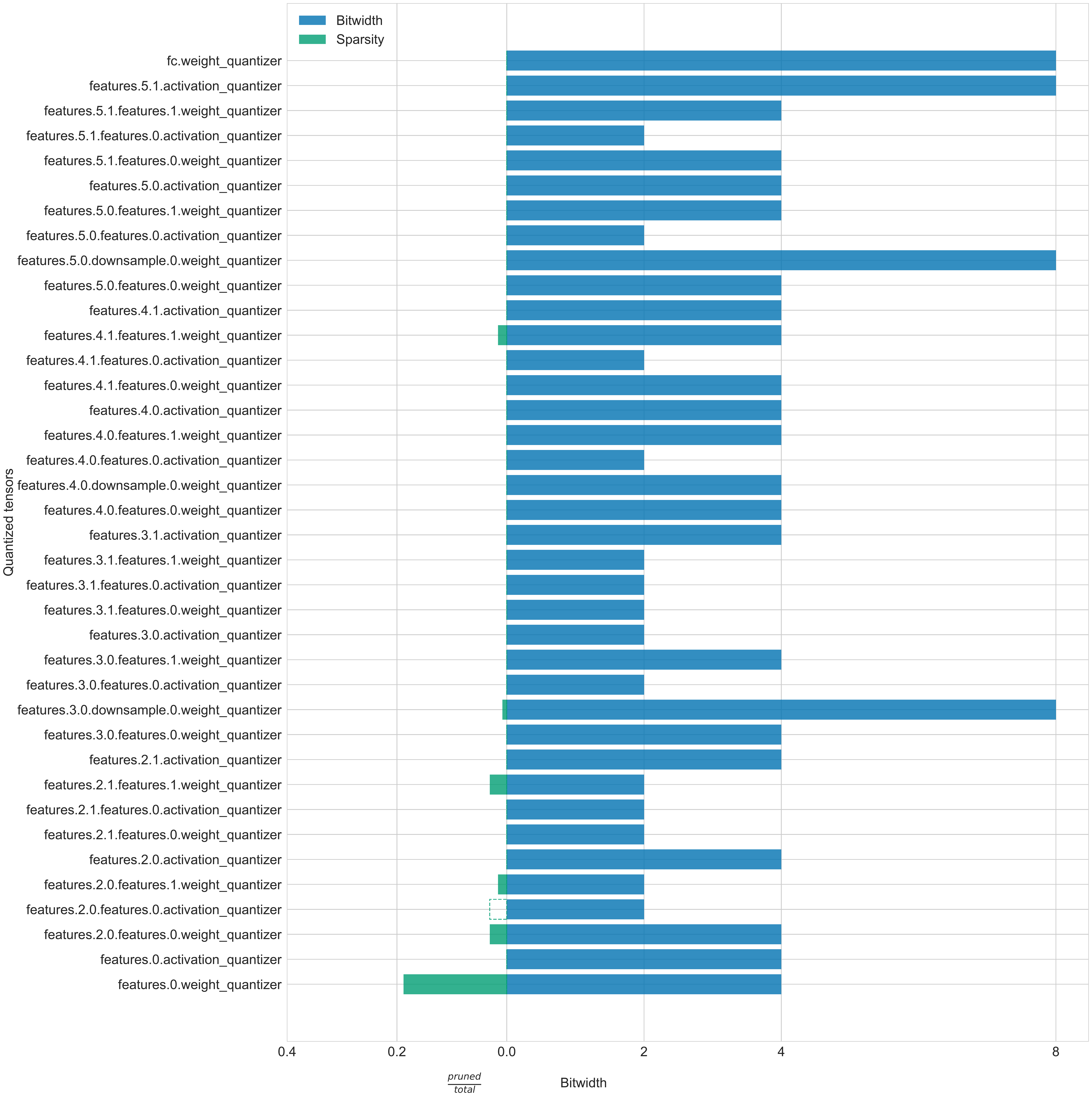}
  \caption{Learned ResNet18 architecture for first run with $\mu = 0.07$.}\label{fig:resnet_007}
\end{figure*}

\begin{figure*}[ht!]
  \centering
  \includegraphics[width=\linewidth]{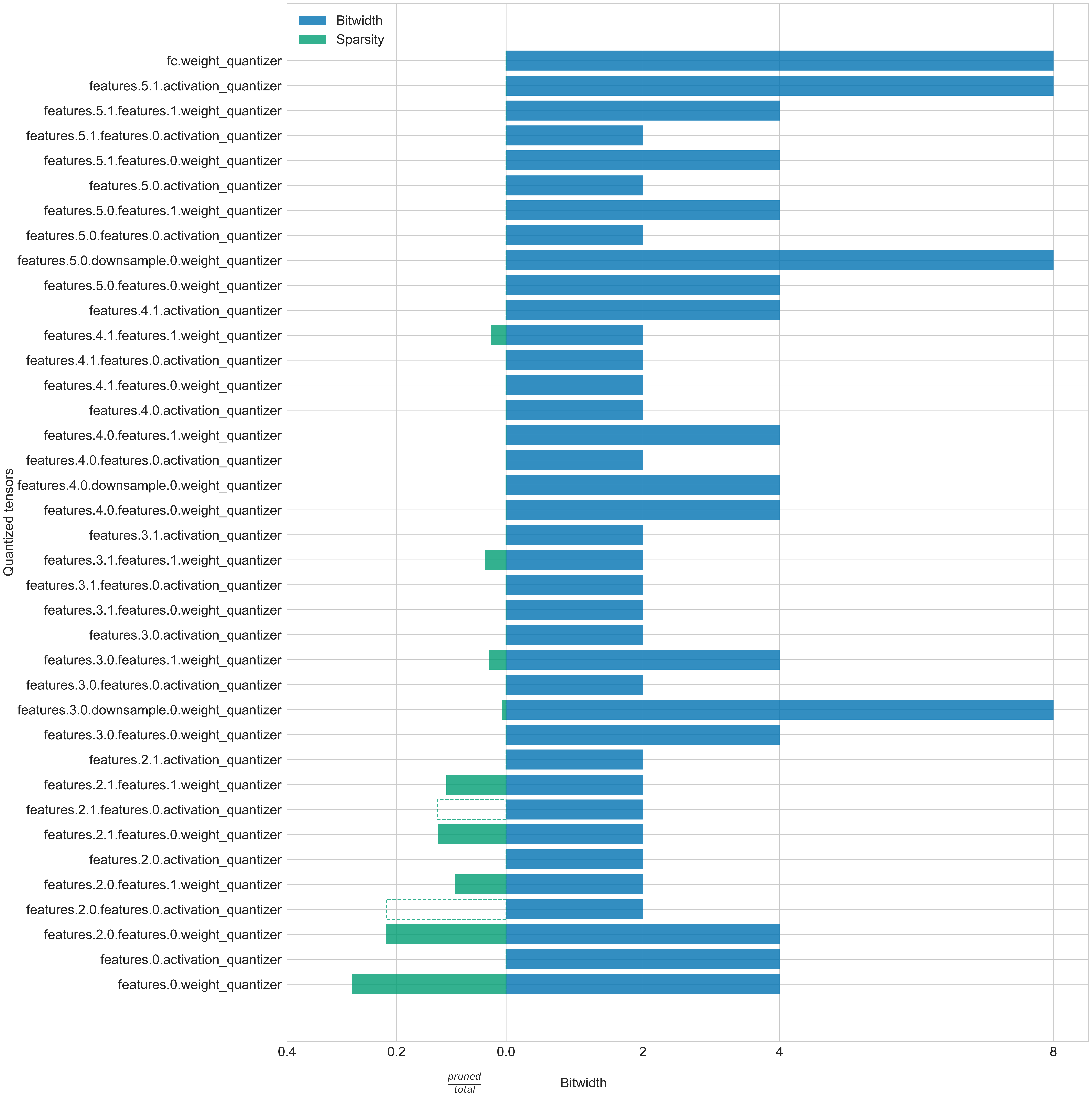}
  \caption{Learned ResNet18 architecture for first run with $\mu = 0.2$.}\label{fig:resnet_02}
\end{figure*}

Full results for Bayesian Bits are provided in table~\ref{tab:res_imagenet}, Figure~\ref{fig:imagenet_scratch_results} the corresponding plot, whereas Figures~\ref{fig:resnet_003},~\ref{fig:resnet_005},~\ref{fig:resnet_007} and~\ref{fig:resnet_02} provide the learned ResNet18 architectures using various regularization strengths. It is interesting to see that the learned architectures tend to have higher bit precision for the first and last layers as well as on the weights that correspond to some of the shortcut connections. 

\begin{table}[ht]
\small
\centering
\caption{Results on the Imagenet task with the ResNet18 architecture. We compare against methods from the previous experiments as well as PACT~\cite{choi2018pact}, \citep{jacob2018quantization} and DQ~\citep{uhlich2019dq}. * indicates first and last layers in full precision. ** first and last layers in 8 bits. NB: for \cite{choi2018pact} these results are hypothetical and based on the assumption that changing the first and last layers in 8 bits does not harm accuracy. LSQ \cite{esser2019lsq} results are run by us. QO, PO48 and PO8 indicate ablation study results. The experiments labeled `Updated' indicate experiments in which activations feeding into residual connections are not quantized. See section D.1 for details.}
\begin{tabular}{lccc}
\toprule
Method & \# bits W/A & Top-1 Acc. (\%) & Rel. GBOPs (\%) \\
\midrule
Full precision & 32/32 & 69.68 & 100 \\
QT \citep{jacob2018quantization} & 8/8 & 70.38 & 6.25 \\
TWN \citep{li2016ternary} & 2/32 & 61.80 & 5.95 \\
LR-Net \citep{shayer2017learning}* & 1/32 & 59.90 & 4.58 \\
RQ \citep{louizos2018relaxed} & 5/5 & 65.10 & 2.54 \\
PACT \citep{choi2018pact} & 4/4* & 69.20 & 3.12 \\
PACT \citep{choi2018pact}  & 2/2* & 64.40 & 2.02 \\
PACT \citep{choi2018pact} & 4/4** & 69.20 & 1.87 \\
PACT \citep{choi2018pact}  & 2/2** & 64.40 & 0.77 \\
LSQ \cite{esser2019lsq} & 8/8 & 70.48 & 6.25 \\
LSQ \cite{esser2019lsq} & 4/8 & 3.13 & 69.7 \\
LSQ \cite{esser2019lsq} & 4/4** & 1.87 & 68.87 \\
LSQ \cite{esser2019lsq} & 4/4 & 1.56 & 68.38 \\
DQ \citep{uhlich2019dq} & Mixed  & 68.40 & 2.10 \\
DQ - restricted \citep{uhlich2019dq} & Mixed & 68.40 & 3.09 \\ 
\midrule

Bayesian Bits $\mu=0.01$ (Pre-FT) & Mixed & 69.70$\pm$0.03 & 2.85$\pm$0.04 \\
Bayesian Bits $\mu=0.01$ & Mixed & 69.84$\pm$0.02 & 2.85$\pm$0.04 \\

Bayesian Bits $\mu=0.03$ (Pre-FT) & Mixed & 69.16$\pm$0.10 & 1.93$\pm$0.05 \\
Bayesian Bits $\mu=0.03$ & Mixed  & 69.39$\pm$0.05 & 1.93$\pm$0.05 \\

Bayesian Bits $\mu=0.05$ (Pre-FT) & Mixed & 67.96$\pm$0.22& 1.44$\pm$0.05 \\
Bayesian Bits $\mu=0.05$ & Mixed & 68.21$\pm$0.23& 1.44$\pm$0.05 \\

Bayesian Bits $\mu=0.07$ (Pre-FT) & Mixed & 66.27$\pm$0.15 & 1.06$\pm$0.02 \\
Bayesian Bits $\mu=0.07$ & Mixed & 66.81$\pm$0.13 & 1.06$\pm$0.02 \\

Bayesian Bits $\mu=0.2$ (Pre-FT) & Mixed & 62.32$\pm$0.71 & 0.68$\pm$0.03\\
Bayesian Bits $\mu=0.2$ & Mixed & 63.76$\pm$0.34 & 0.68$\pm$0.03 \\

\midrule
Bayesian Bits, QO; $\mu=0.01$ & Mixed & 69.85 $\pm$ 0.06 & 3.00 $\pm$ 0.03\\
Bayesian Bits, QO; $\mu=0.03$ & Mixed & 68.80 $\pm$ 0.37 & 1.78 $\pm$ 0.11\\
Bayesian Bits, QO; $\mu=0.05$ & Mixed & 67.70 $\pm$ 0.53 & 1.42 $\pm$ 0.10\\
Bayesian Bits, QO; $\mu=0.07$ & Mixed & 67.59 $\pm$ 0.02 & 1.19 $\pm$ 0.01\\
\midrule
Bayesian Bits, Updated; $\mu=0.01$ & Mixed & 69.82 $\pm$ 0.07 & 2.41 $\pm$ 0.03\\
Bayesian Bits, Updated; $\mu=0.02$ & Mixed & 69.61 $\pm$ 0.08 & 1.99 $\pm$ 0.00\\
Bayesian Bits, Updated; $\mu=0.03$ & Mixed & 69.18 $\pm$ 0.08 & 1.70 $\pm$ 0.03\\
Bayesian Bits, Updated; $\mu=0.05$ & Mixed & 68.17 $\pm$ 0.18 & 1.38 $\pm$ 0.03\\
Bayesian Bits, Updated; $\mu=0.07$ & Mixed & 66.85 $\pm$ 0.17 & 1.07 $\pm$ 0.01\\
Bayesian Bits, Updated; $\mu=0.1$ & Mixed & 65.89 $\pm$ 0.13 & 0.81 $\pm$ 0.00\\
Bayesian Bits, Updated; $\mu=0.2$ & Mixed & 62.73 $\pm$ 0.10 & 0.52 $\pm$ 0.00\\
\midrule
Bayesian Bits, Updated, QO; $\mu=0.01$ & Mixed & 69.88 $\pm$ 0.03 & 2.51 $\pm$ 0.04\\
Bayesian Bits, Updated, QO; $\mu=0.02$ & Mixed & 69.62 $\pm$ 0.04 & 2.11 $\pm$ 0.05\\
Bayesian Bits, Updated, QO; $\mu=0.03$ & Mixed & 69.16 $\pm$ 0.11 & 1.80 $\pm$ 0.01\\
Bayesian Bits, Updated, QO; $\mu=0.05$ & Mixed & 68.21 $\pm$ 0.07 & 1.44 $\pm$ 0.01\\
Bayesian Bits, Updated, QO; $\mu=0.07$ & Mixed & 67.09 $\pm$ 0.16 & 1.02 $\pm$ 0.01\\
Bayesian Bits, Updated, QO; $\mu=0.2$ & Mixed & 63.56 $\pm$ 0.17 & 0.62 $\pm$ 0.00\\
\midrule
Bayesian Bits, PO48; $\mu=0.01$ & Mixed & 69.79 $\pm$ 0.02 & 3.10 $\pm$ 0.00\\
Bayesian Bits, PO48; $\mu=0.2$ & Mixed & 69.69 $\pm$ 0.04 & 2.86 $\pm$ 0.01\\
Bayesian Bits, PO48; $\mu=0.5$ & Mixed & 67.72 $\pm$ 0.05 & 1.86 $\pm$ 0.00\\
Bayesian Bits, PO48; $\mu=0.7$ & Mixed & 66.08 $\pm$ 0.01 & 1.51 $\pm$ 0.00\\
Bayesian Bits, PO48; $\mu=1.0$ & Mixed & 63.09 $\pm$ 0.06 & 1.17 $\pm$ 0.00\\
\midrule
Bayesian Bits, PO8; $\mu=0.01$ & Mixed & 70.54 $\pm$ 0.02 & 6.20 $\pm$ 0.00\\
Bayesian Bits, PO8; $\mu=0.2$ & Mixed & 70.28 $\pm$ 0.05 & 5.65 $\pm$ 0.01\\
Bayesian Bits, PO8; $\mu=0.5$ & Mixed & 68.37 $\pm$ 0.05 & 3.68 $\pm$ 0.01\\
Bayesian Bits, PO8; $\mu=0.7$ & Mixed & 67.02 $\pm$ 0.02 & 3.00 $\pm$ 0.01\\
Bayesian Bits, PO8; $\mu=1.0$ & Mixed & 64.34 $\pm$ 0.05 & 2.29 $\pm$ 0.01\\

\bottomrule
\end{tabular}
\label{tab:res_imagenet}
\end{table}

\subsubsection{ImageNet ResNet18 gate evolution}
\begin{figure}[h]
  \centering
  \includegraphics[width=0.9\linewidth]{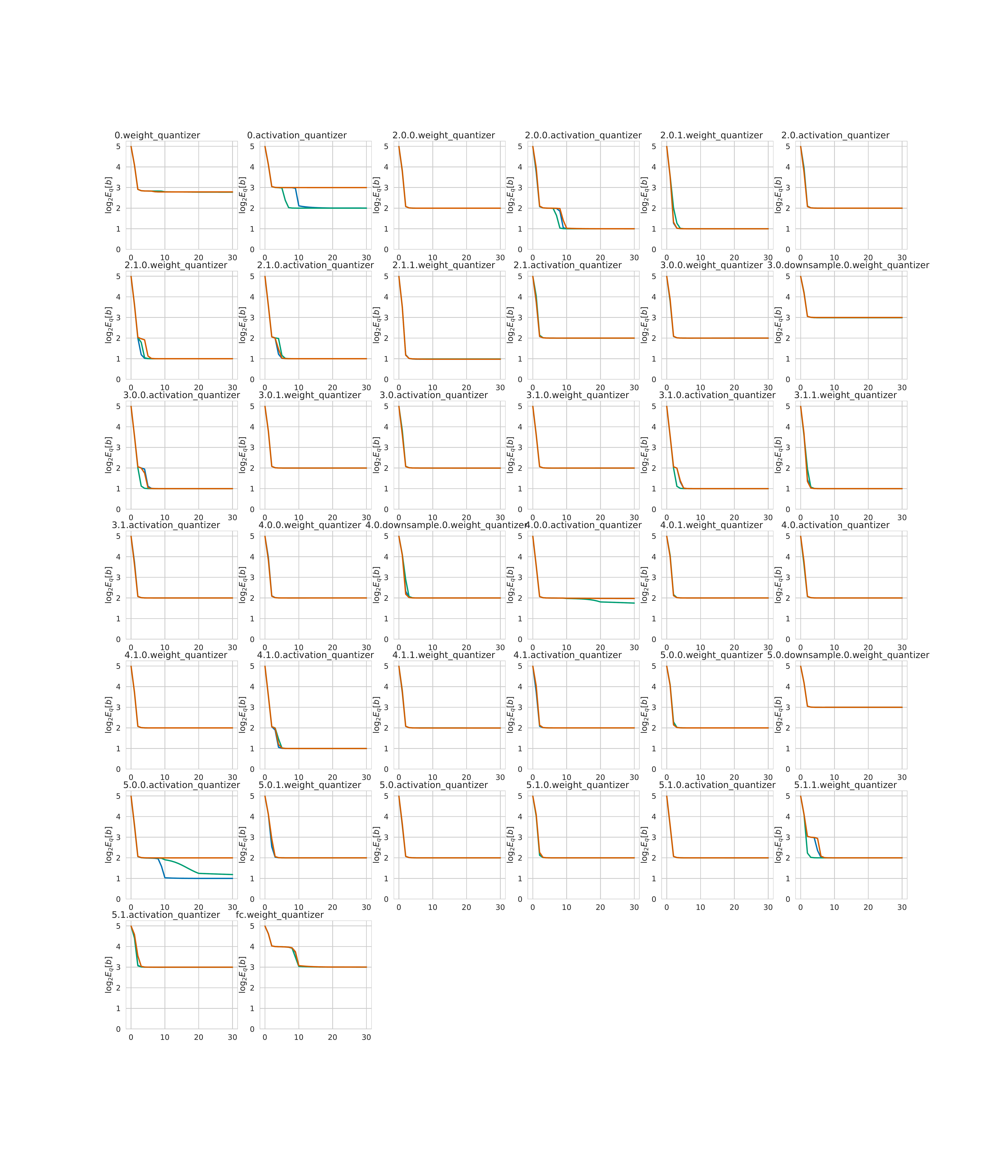}
  \caption{The evolution of gates for three ResNet18 ImageNet experiments with $\mu=0.05$}\label{fig:gate-evol}
\end{figure}

The evolution of the gates for three experiments, with $\mu=0.05$ are plotted in Figure~\ref{fig:gate-evol}.

\subsubsection{ImageNet post-training}
\begin{figure}[h]
  \centering
  \includegraphics[width=0.9\linewidth]{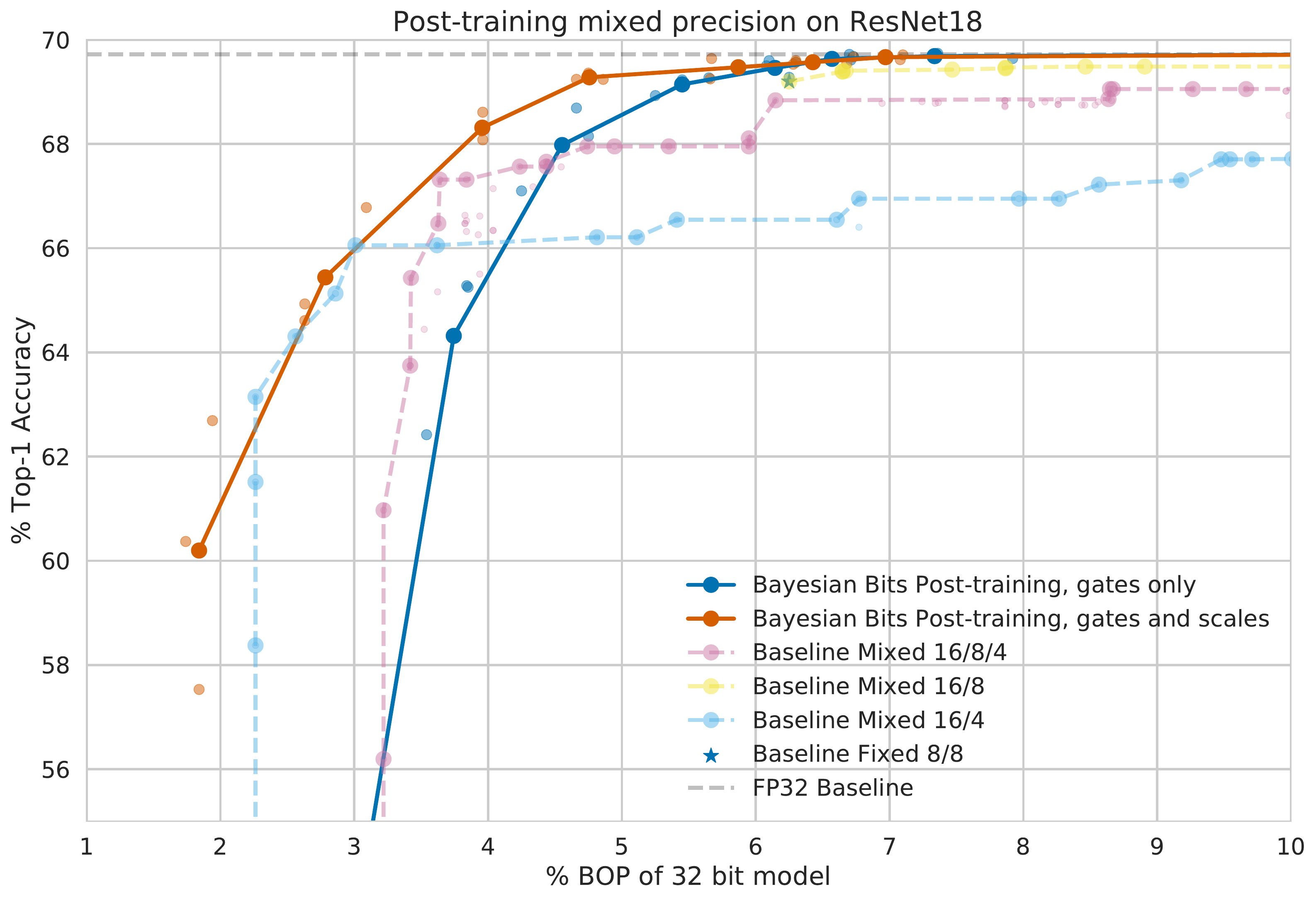}
  \caption{Bayesian Bits Imagenet validation accuracy on ResNet18 for $\mu\in\{0.03, 0.05, 0.07, 0.2\}$. Means and individual runs of 3 training runs for each $\mu$.}\label{fig:post_results_plot}
\end{figure}

Full results from the post-training quantization experiment are provided in Table~\ref{tab:post-training} as well as in the updated plot in Figure~\ref{fig:post_results_plot}. In the baseline experiment we first measured quantization sensitivity for each quantizer in the network by keeping the network in INT16, while setting the target quantizer to a lower bit-width. We then sorted all quantizers in order of increasing sensitivity, and set the quantizer to the lower bit width cumulatively, measuring the accuracy after each step. Figure~\ref{fig:imagenet_pre_post_ft} shows the Pareto front of these results, as results do not monotonically decrease with more quantizers turned on.

\begin{table}[]
    \centering
    \begin{tabular}{lcccc}
        \toprule %
         & \multicolumn{2}{c}{Gates only} & \multicolumn{2}{c}{Gates and scales} \\
        Regularization & Top-1 Acc. (\%) & Rel. GBOPs (\%) & Top-1 Acc. (\%) & Rel. GBOPs (\%) \\
        \midrule
        $\mu=0.0001$ & 69.73 $\pm$ 0.06 & 12.05 $\pm$ 0.68 & 69.72 $\pm$ 0.05 & 10.87 $\pm$ 0.40\\
        $\mu=0.0005$ & 69.69 $\pm$ 0.03 & 7.34 $\pm$ 0.34 & 69.67 $\pm$ 0.03 & 6.97 $\pm$ 0.12\\
        $\mu=0.001$ & 69.63 $\pm$ 0.04 & 6.57 $\pm$ 0.14 & 69.57 $\pm$ 0.02 & 6.43 $\pm$ 0.13\\
        $\mu=0.0025$ & 69.46 $\pm$ 0.09 & 6.14 $\pm$ 0.05 & 69.47 $\pm$ 0.12 & 5.87 $\pm$ 0.21\\
        $\mu=0.005$ & 69.14 $\pm$ 0.11 & 5.45 $\pm$ 0.12  & 69.28 $\pm$ 0.04 & 4.76 $\pm$ 0.06\\
        $\mu=0.01$ & 67.98 $\pm$ 0.47 & 4.55 $\pm$ 0.15 & 68.31 $\pm$ 0.16 & 3.96 $\pm$ 0.00 \\
        $\mu=0.02$ & 64.32 $\pm$ 0.95 & 3.74 $\pm$ 0.10 & 65.44 $\pm$ 0.68 & 2.78 $\pm$ 0.15\\
        $\mu=0.05$ & 51.31 $\pm$ 1.93 & 2.90 $\pm$ 0.02 &  60.20 $\pm$ 1.49 & 1.84 $\pm$ 0.06\\
        \bottomrule
    \end{tabular}
    \caption{Results on learning only the gates (left) and both the gates and the scales (right) on a small dataset for various regularization strengths. Means and standard errors are computed over 3 training runs for each value of $\mu$.}
    \label{tab:post-training}
\end{table}

\begin{figure*}
    \centering
    \includegraphics[width=0.8\linewidth]{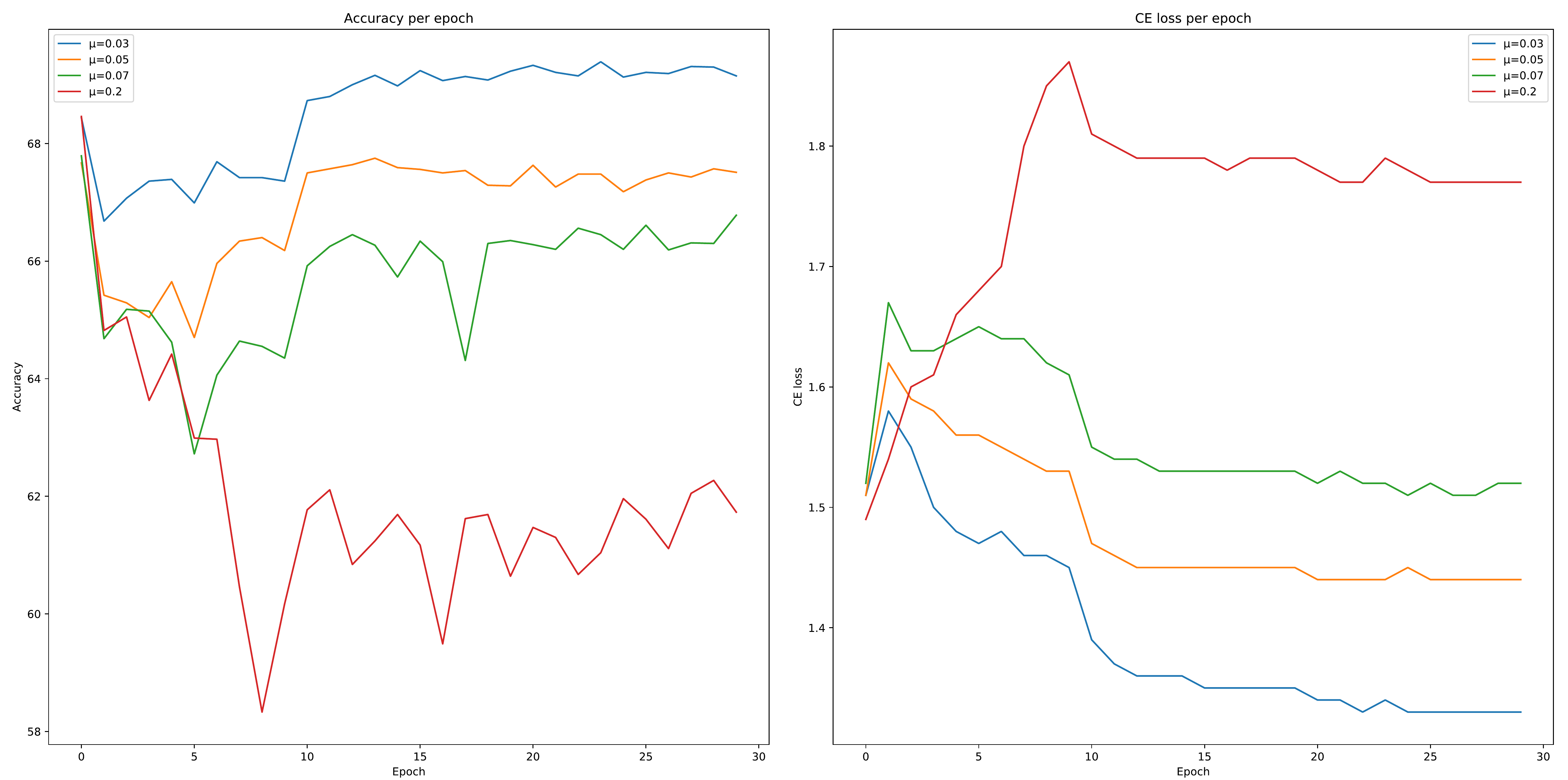}
    \caption{Evolution of validation accuracy and (per epoch average) cross-entropy loss during training of ResNet18 on ImageNet, first run for $\mu\in\{0.03, 0.05, 0.07, 0.2\}$}
\end{figure*}

\begin{figure*}
    \centering
    \includegraphics[width=0.8\linewidth]{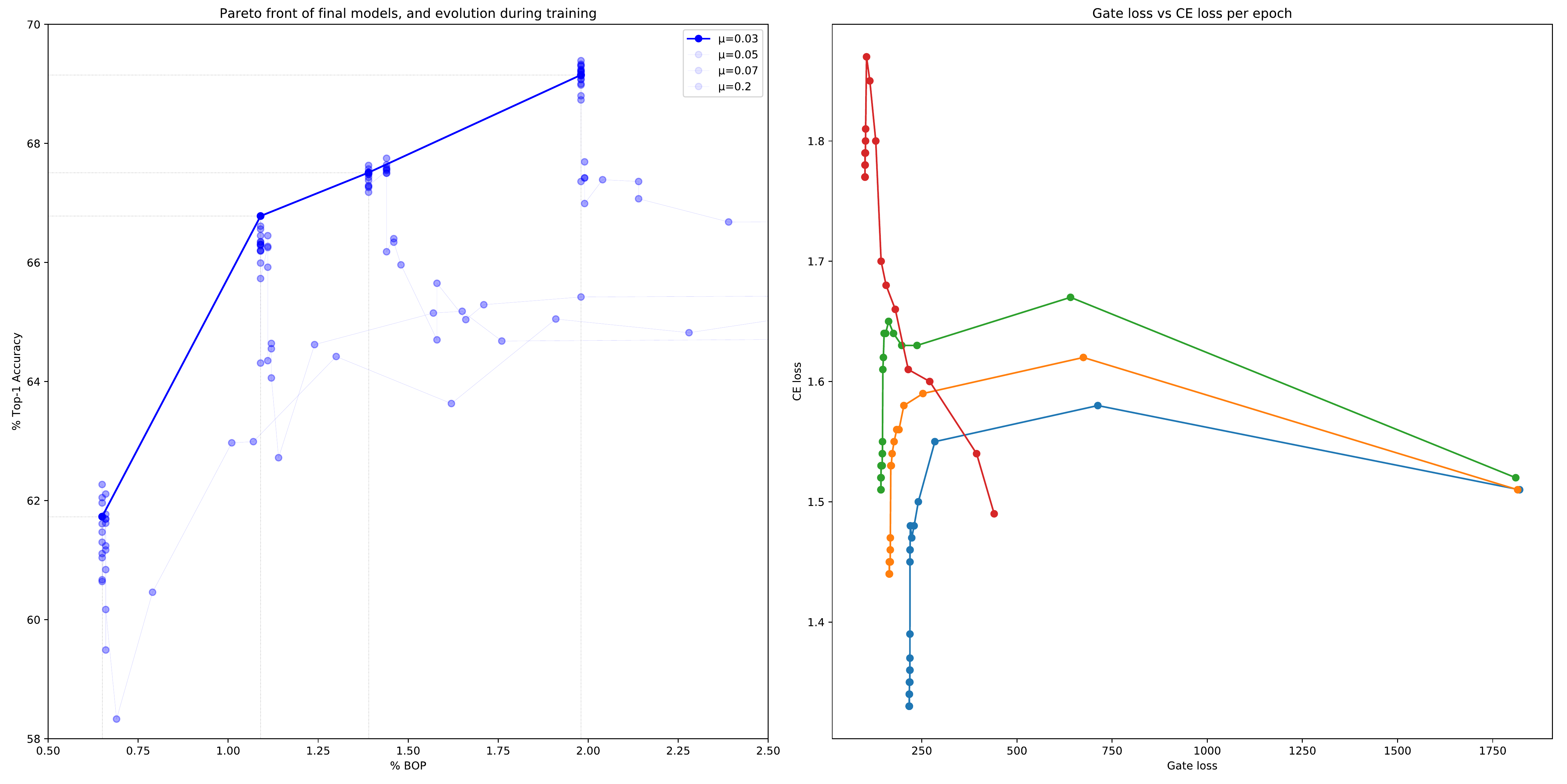}
    \caption{Left plot: Pareto front of final model efficiency vs accuracy trade-offs, including evolution towards final trade-offs. Right plot: Co-evolution of cross-entropy and gate loss per epoch. Both plots show results of training ResNet18 on Imagenet, first run for $\mu\in\{0.03, 0.05, 0.07, 0.2\}$}
\end{figure*}

\begin{figure*}
    \centering
    \includegraphics[width=0.8\linewidth]{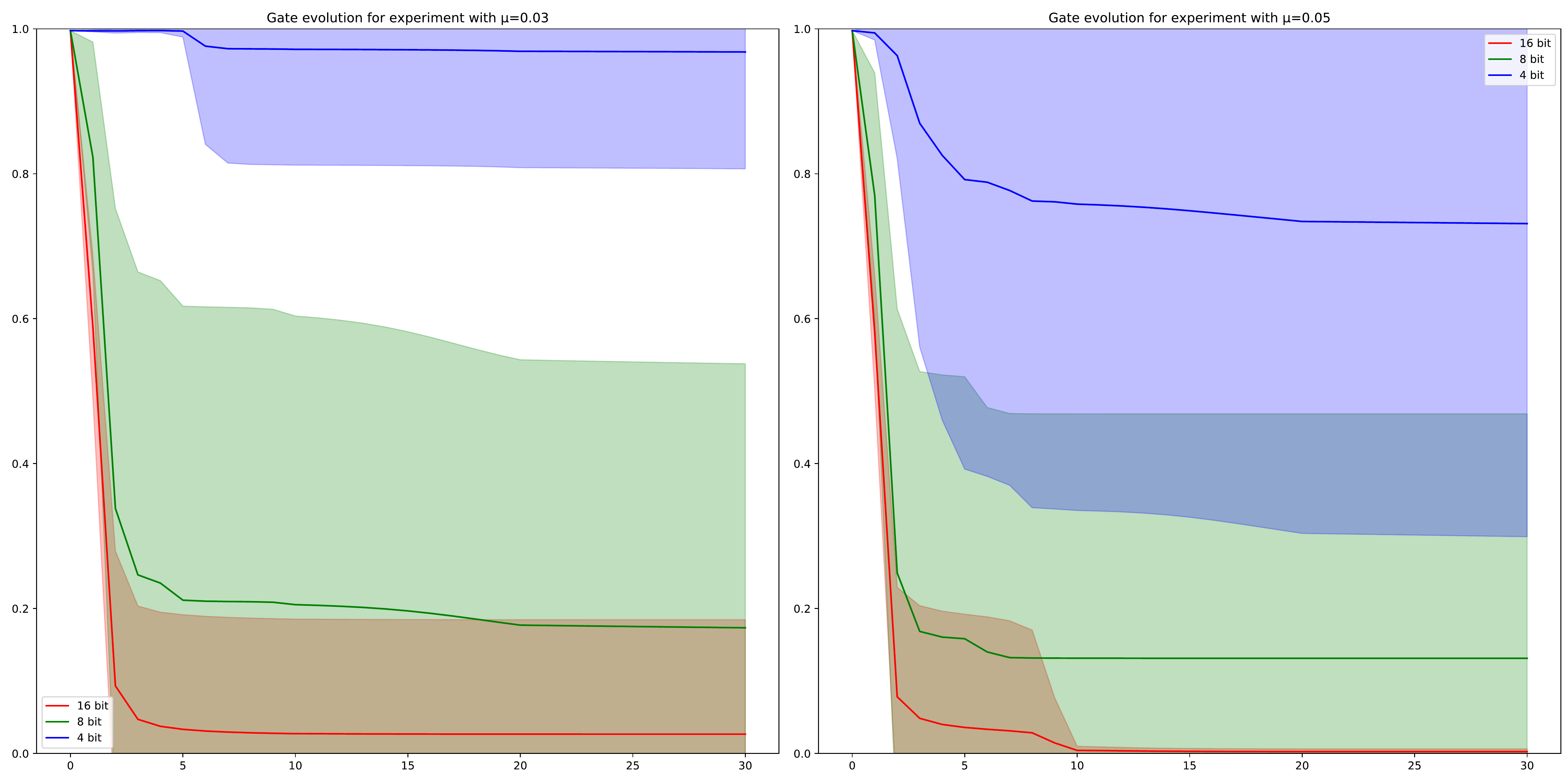}
    \caption{Evolution of training of ResNet18 ImageNet experiments, first run for $\mu\in\{0.03, 0.05\}$. Mean gate probability with shaded area indicating 1 standard deviation.}
\end{figure*}

\begin{figure*}
    \centering
    \includegraphics[width=0.8\linewidth]{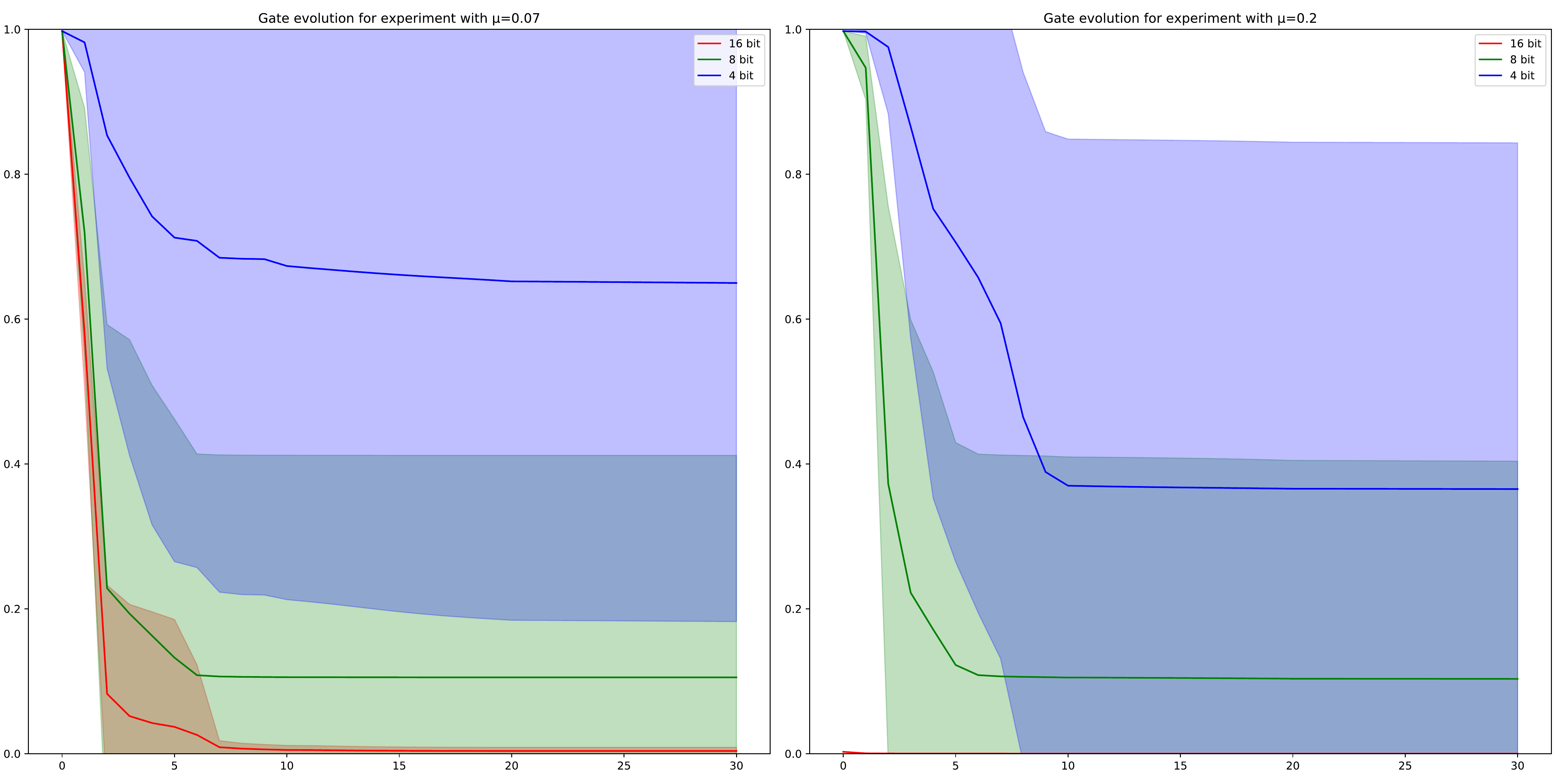}
    \caption{Evolution of training of ResNet18 ImageNet experiments, first run for $\mu\in\{0.07, 0.2\}$. Mean gate probability with shaded area indicating 1 standard deviation.}
\end{figure*}

\end{document}